\documentclass[10pt,journal,compsoc]{IEEEtran}
\usepackage{times}
\usepackage{epsfig}
\usepackage{graphicx}
\usepackage{amsmath}
\usepackage{amssymb}
\usepackage{amsthm}
\usepackage{bm}
\usepackage{subfigure}
\usepackage[linesnumbered, ruled]{algorithm2e}
\usepackage{booktabs}
\usepackage{multirow}
\usepackage{array}
\usepackage{enumitem}
\usepackage{float}
\usepackage{color}
\usepackage{url}
\usepackage{hanging}
\usepackage{algorithmic}
\usepackage{geometry}

\graphicspath{{../}}
\newcommand{\red}[1]{{\color{red}#1}}
\newcommand{\blue}[1]{{\color{blue}#1}}

\newcommand{\modify}[1]{{\color{black}#1}}
\newcommand{\re}[1]{{\color{black}#1}}
\newcommand{\rere}[1]{{\color{black}#1}}
\newcommand{\erhao}{\fontsize{21pt}{\baselineskip}\selectfont}

\begin{document}
\newgeometry{top=6cm,bottom=1cm}

\onecolumn{

\noindent \textbf{\erhao{Unsupervised Person Re-identification by Deep Asymmetric Metric Embedding}}

\vspace{2cm}

\noindent {\LARGE{Hong-Xing Yu, Ancong Wu, Wei-Shi Zheng}}

\Large
\vspace{2cm}

\noindent Demo code is available at: \\
\ \ \ \ \ \ \ \ \ \ \ \ \url{https://github.com/KovenYu/DECAMEL}

\vspace{1cm}

\noindent For reference of this work, please cite:

\vspace{1cm}
\noindent Hong-Xing Yu, Ancong Wu, Wei-Shi Zheng.
``Unsupervised Person Re-identification by Deep Asymmetric Metric Embedding.''
\emph{IEEE Transactions on Pattern Analysis and Machine Intelligence.} (DOI 10.1109/TPAMI.2018.2886878).

\vspace{1cm}

\noindent Bib:\\
\noindent
@article\{yu2018unsupervised,\\
\ \ \   title=\{Unsupervised Person Re-identification by Deep Asymmetric Metric Embedding\},\\
\ \ \  author=\{Yu, Hong-Xing and Wu, Ancong and Zheng, Wei-Shi\},\\
\ \ \  journal=\{IEEE Transactions on Pattern Analysis and Machine Intelligence\\(DOI 10.1109/TPAMI.2018.2886878)\},\\
\}

}

\clearpage

\newpage
\restoregeometry
%
\title{Unsupervised Person Re-identification by \\ Deep Asymmetric Metric Embedding}

\author{Hong-Xing Yu, Ancong Wu, Wei-Shi Zheng
\IEEEcompsocitemizethanks{
\IEEEcompsocthanksitem Hong-Xing Yu is with School of Data and Computer Science,
	Sun Yat-sen University, Guangzhou 510275, China, and is also with the Guangdong Province Key Laboratory of Information Security, China.
	{\em E-mail}: xKoven@gmail.com.
\IEEEcompsocthanksitem Ancong Wu is with
	School of Electronics and Information Technology,
	Sun Yat-sen University, Guangzhou 510275, China, and is also with the Collaborative Innovation Center of High Performance Computing, NUDT, China. {\em E-mail}: wuancong@mail2.sysu.edu.cn.
\IEEEcompsocthanksitem Wei-Shi Zheng is with
	School of Data and Computer Science,
	Sun Yat-sen University,  Guangzhou 510275, China, and is also with the Key Laboratory of Machine Intelligence and Advanced Computing, Ministry of Education, China
	{\em E-mail}: wszheng@ieee.org / zhwshi@mail.sysu.edu.cn.
	\protect\\ (Corresponding author: Wei-Shi Zheng.)
}}

%
%

\markboth{Citation information: DOI 10.1109/TPAMI.2018.2886878, IEEE
Transactions on Pattern Analysis and Machine Intelligence}%
{Shell \MakeLowercase{\textit{et al.}}: Submission to IEEE TPAMI}
%



\IEEEtitleabstractindextext{%
\begin{abstract}
Person re-identification (Re-ID) aims to match identities across non-overlapping camera views.
Researchers have proposed many supervised Re-ID models which require quantities of cross-view pairwise labelled data.
This limits their scalabilities to many applications where a large amount of data from multiple disjoint camera views is available but unlabelled.
Although some unsupervised Re-ID models have been proposed to address the scalability problem,
they often suffer from the view-specific bias problem which is caused by dramatic variances across different camera views,
e.g., different illumination, viewpoints and occlusion.
The dramatic variances induce specific feature distortions in different camera views,
which can be very disturbing in finding cross-view discriminative information for Re-ID in the unsupervised scenarios,
since no label information is available to help alleviate the bias.
We propose to explicitly address this problem by learning an unsupervised asymmetric distance metric based on cross-view clustering.
The asymmetric distance metric allows specific feature transformations for each camera view to tackle the specific feature distortions.
We then design a novel unsupervised loss function to embed the asymmetric metric into a deep neural network, and therefore develop a novel unsupervised deep framework named the \textbf{DE}ep \textbf{C}lustering-based \textbf{A}symmetric \textbf{ME}tric \textbf{L}earning (\emph{DECAMEL}).
In such a way, DECAMEL jointly learns the feature representation and the unsupervised asymmetric metric.
DECAMEL learns a compact cross-view cluster structure of Re-ID data,
and thus help alleviate the view-specific bias and facilitate mining the potential cross-view discriminative information for unsupervised Re-ID.
Extensive experiments on seven benchmark datasets whose sizes span several orders
show the effectiveness of our framework.
\end{abstract}

}

\maketitle

\IEEEdisplaynontitleabstractindextext

%
\IEEEpeerreviewmaketitle

\IEEEraisesectionheading{\section{Introduction}\label{sec:introduction}}

%
%
%
%

\begin{figure*}[!t]
\centering
\subfigure[View-specific conditions]{
\includegraphics[width=0.273\linewidth, height=0.35\linewidth]{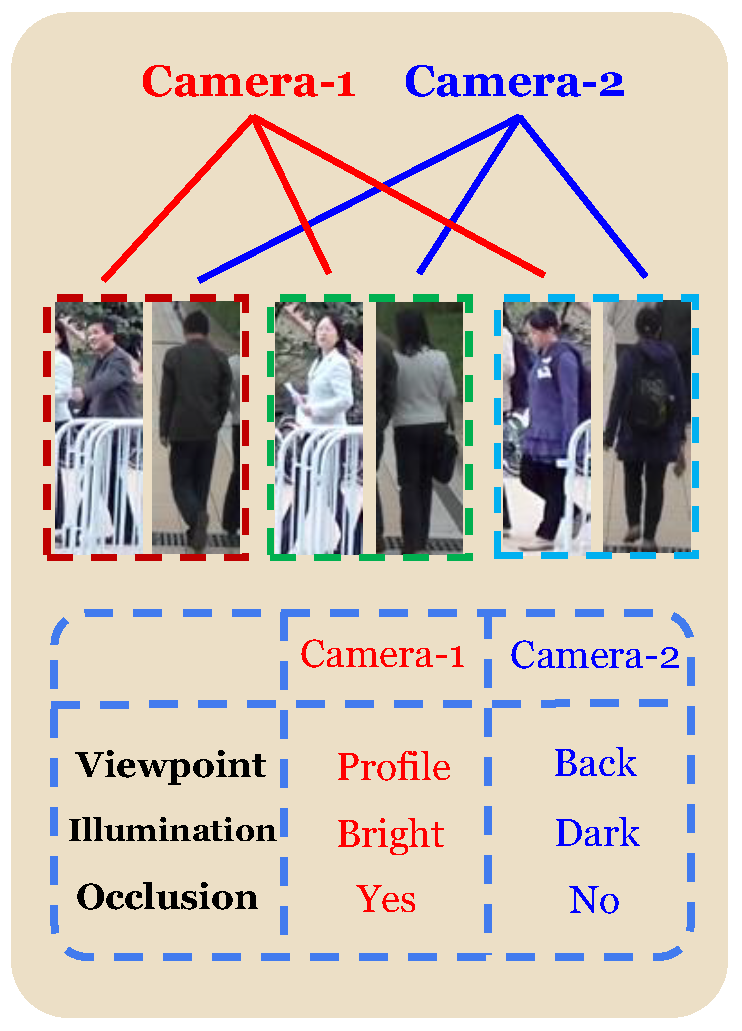}\label{figure:viewingConditionVariation}
}
\subfigure[View-specific bias in feature space]{
\includegraphics[width=0.375\linewidth, height=0.35\linewidth]{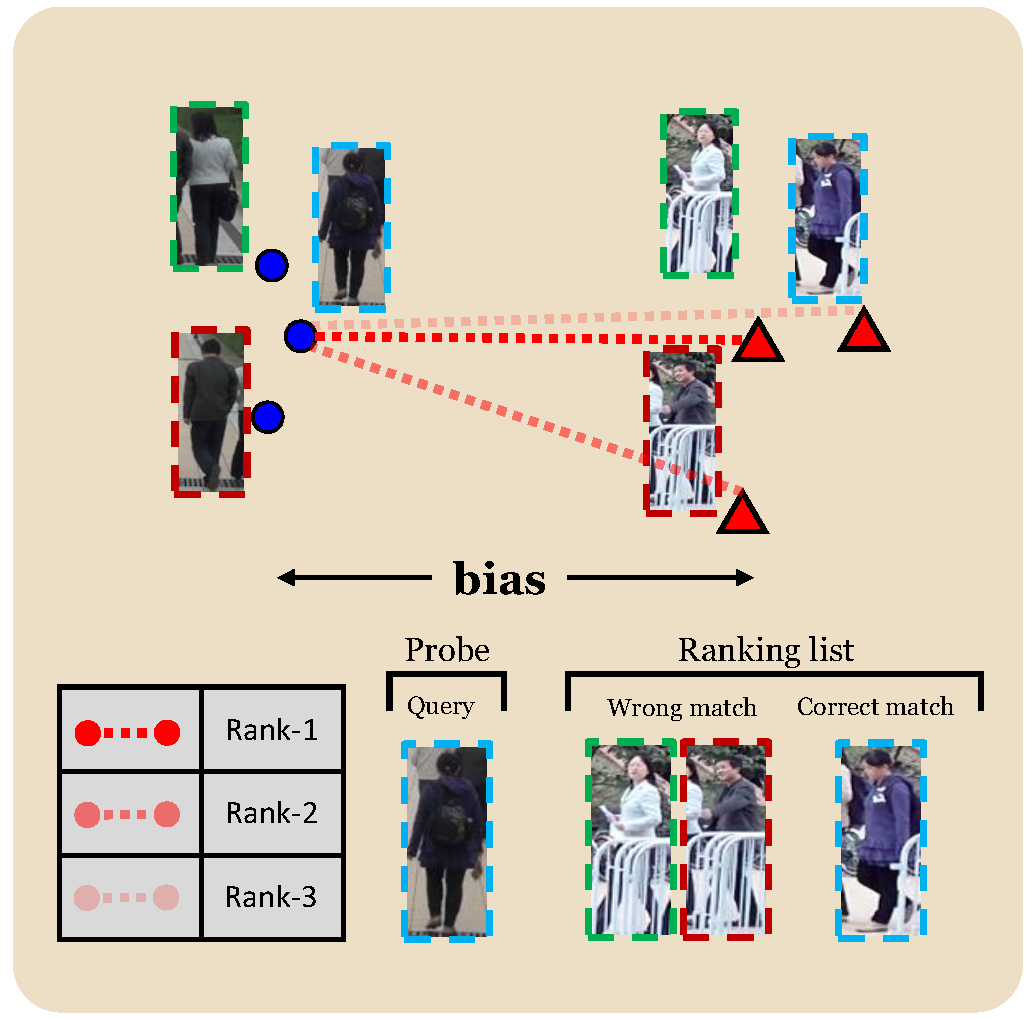}\label{figure:viewSpecificBias}
}
\subfigure[View-specific bias alleviated]{
\includegraphics[width=0.273\linewidth, height=0.35\linewidth]{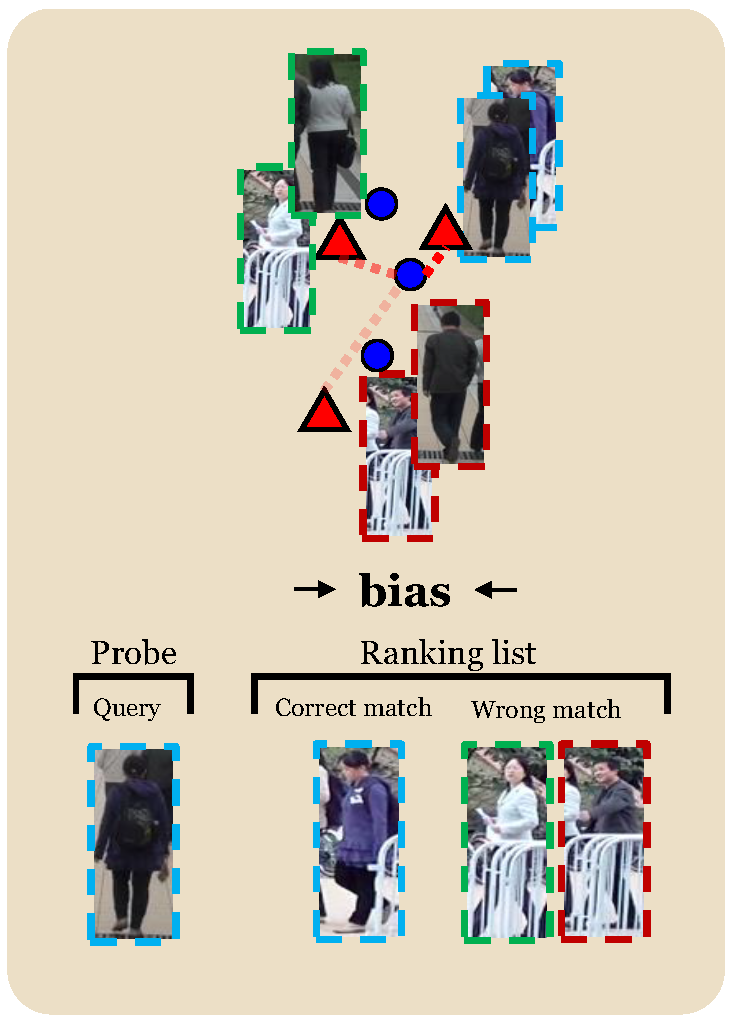}\label{figure:viewSpecificBiasAlleviated}
}
\vspace{-0.3cm}
\caption{\label{figure:Figure1}
Cause and effect of the view-specific bias.
\textbf{\emph{(a)}} The images from different camera views suffer from
dramatic viewing condition variations across camera views.
Images in the same colored box belong to the same person.
\textbf{\emph{(b)}} The specific viewing conditions
lead to the view-specific feature distortions/bias,
making unsupervised Re-ID more challenging.
\re{A visualization of the view-specific bias can be found in Figure \ref{figure:originalDistribution}}.
For example, given a probe image (in blue box), the correct match falls to the third rank due to the view-specific bias.
\textbf{\emph{(c)}} If we can tackle the view-specific feature distortion (i.e., alleviate the view-specific bias), we may reach a better cross-view matching performance.
\re{Further visualization and discussion can be found in Figure \ref{figure:distribution} and Figure \ref{figure:FM}.}
(Best viewed in color).
}
\vspace{-0.5cm}
\end{figure*}

Along with the extensive deployment of visual surveillance networks,
considerable visual surveillance data is emerging everyday within the networks.
A basic problem of analyzing and exploiting the data
is to find target persons who had been previously observed like missing children and suspects,
also known as person re-identification (Re-ID) \cite{bedagkar2014survey, 2013_ACM_survey}.
Typically, Re-ID focuses on pedestrian matching and ranking across multiple non-overlapping camera views.


Re-ID remains an open problem although it has received increasing explorations in recent years,
principally because of dramatic intra-person appearance variation across views and high inter-person similarity.
They mainly focus on learning robust and discriminative feature representations
\cite{f2,f3,f4,f5,f6,f8,f9,f15,f16,f17,f18} and
distance metrics
\cite{m1,m3,m4,m5,m6,m7,m8,m9,m15,m16,m18,m20,m22,m25,2015_TCSVT_ASM}.
Recently, deep learning has been adopted to the Re-ID community and has achieved promising performances
\cite{d1,d2,d3,d4,d5,d6,d7,d8,d9,d10,d13,d14}.
They have contributed a lot to the Re-ID community.

However, supervised models are intrinsically limited because they rely on a large amount of correctly labelled cross-view training data,
which is very expensive \cite{2015_TCSVT_xiaojuan}.
In the context of Re-ID, this limitation is even pronounced because \emph{(1)} manual labelling may not be fully reliable when a huge number of images to be checked across multiple camera views,
and more importantly \emph{(2)} the astronomical cost of time and money is prohibitive to label the overwhelming amount of data.
In many scenarios,
there is a large amount of data available but unlabelled,
so that the supervised methods would be restricted or not applicable.

To directly make full use of the cheap and valuable unlabelled data,
some efforts on unsupervised Re-ID have been made \cite{SDALF,CPS,SDC,GTS,ISR,RKSL,Dic,UDML,GL},
\re{which typically learn general/universal feature projections for person images from every camera view,
but the performances are still not satisfactory.
One of the main reasons is that without the supervision of cross-view pairwise labelled data,
it is very difficult for a universal feature projection to capture the cross-view discriminative information under dramatic cross-view person appearance variations
caused by view-specific conditions (Figure \ref{figure:viewingConditionVariation}).
For example, a person in white may appear wearing gray in one camera view where illumination is darker,
while he may appear snow-white in another view where illumination is brighter.
Without the pairwise supervision guidance,
it is very hard for a universal feature projection to map such drastically different cross-view image features of the same person to very close points in the subspace.
}
More generally, the view-specific conditions introduce the view-specific bias, i.e., some specific feature distortions in different camera views,
which can be very disturbing in finding what is more distinguishable in matching people across views.
We show a toy example to illustrate this disturbing effect in Figure \ref{figure:Figure1}.
In Figure \ref{figure:viewingConditionVariation}, the color feature of a person's arm may be located in the central position of images from Camera-1
since Camera-1 typically captures the profile of a person,
while the \modify{corresponding} color feature may appear at the boundary of images from Camera-2 since it captures the back of the person.
Thus, the view-specific feature distortion can make the cross-view matching even harder as shown in Figure \ref{figure:viewSpecificBias}.
In particular, most existing unsupervised models treat the samples from different views in the same manner,
and thus could suffer from the effect of the view-specific bias.

In this work, we propose to \emph{explicitly} deal with the view-specific bias problem in unsupervised Re-ID by formulating it as an unsupervised asymmetric distance metric learning problem.
We briefly introduce the idea in the following.
Given a pair of sample feature representations ${\mathbf{x}_i}$ and ${\mathbf{x}_j}$,
a conventionally learned distance metric between them is:
\begin{equation}\label{equation:symmetric}
\small
d_l(\mathbf{x}_i,\mathbf{x}_j)
= \sqrt{(\mathbf{x}_i-\mathbf{x}_j)^{\mathrm{T}}\bm{M}(\mathbf{x}_i-\mathbf{x}_j)}
= \lVert \bm{U}^{\mathrm{T}}\mathbf{x}_i - \bm{U}^{\mathrm{T}}\mathbf{x}_j \rVert_2,
\end{equation}
where $\bm{M} = \bm{U}\bm{U}^{\mathrm{T}}$ is a positive semi-defined matrix, and $\bm{U}$ is a transformation matrix.
Learning such a metric is equivalent to finding a space ``shared'' by samples from each camera view \cite{kulis2013metric}.
This shared space is found by projecting all samples with a view-generic universal transformation $\bm{U}$.
However, different camera views may induce different feature characteristics,
e.g., the side-view of persons in Camera-1 vs. the back-view in Camera-2 as shown in Figure \ref{figure:viewingConditionVariation}.
Intuitively, it is important to perform view-specific transformations for acquiring common features to match person images across camera views
(e.g., selecting the \modify{corresponding} color features at different locations of images from different camera views).
Therefore, it can be hard for a universal transformation to implicitly tackle the view-specific feature distortions from different camera views,
especially when we lack label information to guide it.
This motivates us to explicitly take the view-specific feature distortion into account.
Inspired by the supervised asymmetric distance model \cite{2015_TCSVT_ASM, f2},
we propose to embed the asymmetric metric learning into our unsupervised Re-ID modelling,
and thus consider the modification of Eq. (\ref{equation:symmetric}), i.e., the asymmetric form:
\begin{equation}\label{equation:asymmetric}
\small
d_l(\{\mathbf{x}_i, v_i\}, \{\mathbf{x}_j, v_j\}) = \lVert \bm{U}_{v_i}^\mathrm{T}\mathbf{x}_i - \bm{U}_{v_j}^\mathrm{T}\mathbf{x}_j \rVert_2,
\end{equation}
where $v_i$ denotes which camera view the $i$-th sample comes from,
and $\bm{U}_{v_i}$ is the corresponding view-specific transformation.
Such an asymmetric metric allows specific transformation for each view
to tackle the view-specific feature distortions.

Since no label information is provided to strictly distinguish every visually similar person in unsupervised Re-ID scenarios,
we encourage the asymmetric metric to condense the visually similar cross-view person image \emph{clusters}, and thus better distinguish them from other dissimilar clusters.
With the asymmetric metric in the clustering procedure, we can explicitly address the view-specific bias and learn a better cross-view cluster structure of the Re-ID data in the shared space.

In the following, we refer to the clustering procedure which uses an asymmetric metric as the \emph{asymmetric metric clustering}.
Based on the asymmetric metric clustering, we first develop a linear metric learning model named
\textbf{C}lustering-based \textbf{A}symmetric \textbf{ME}tric \textbf{L}earning (\emph{CAMEL}).
CAMEL jointly learns an asymmetric metric and a cluster separation.
Then, based on CAMEL, we further propose a novel unsupervised deep framework named \textbf{DE}ep \textbf{CAMEL} (\emph{DECAMEL}),
which jointly learns the feature representation and the unsupervised asymmetric metric end to end.
DECAMEL can address the sub-optimality due to the separation of feature and metric learning.
By learning a better cross-view cluster structure in the shared space,
DECAMEL attempts to mine the underlying cross-view discriminative information
to achieve a better cross-view matching performance.

More specifically, DECAMEL consists of a feature extractor network and an asymmetric metric layer.
We propose a novel unsupervised loss function for the whole deep framework.
In the optimization of DECAMEL, the asymmetric metric layer is initialized by CAMEL,
which can alleviate the view-specific bias and learn a preliminary cross-view cluster separation.
Then, the asymmetric metric is embedded into the whole network by the joint learning.
The term ``embedded'' refers to the fact that,
during the joint learning,
the asymmetric metric is back-propagated to the whole network,
so that finally the view-specific bias in the feature space is also alleviated.
This is empirically observed in Figure \ref{figure:FM}.
Through this joint learning procedure,
DECAMEL learns a compact cross-view cluster structure of Re-ID data.
And thereby, it attempts to mine the potential cross-view discriminative information
which will be qualitatively illustrated in Sec. \ref{section:insight} and quantitatively validated in Sec. \ref{section:experiments}.


In summary, our \textbf{main contributions} in this work are:

\re{
\emph{\textbf{(1)}}
We formulate unsupervised Re-ID as a joint learning problem which consists of learning the feature representation and an asymmetric distance metric, together with a cluster separation.
To our best knowledge, this is the first work
formulating unsupervised Re-ID as a joint learning problem of the feature and the metric.
}

\re{
\emph{\textbf{(2)}}
We propose a novel unsupervised deep framework named DECAMEL.
Different from previous works in unsupervised Re-ID,
DECAMEL jointly learns the feature representation and the unsupervised asymmetric metric.
DECAMEL can learn a compact cross-view clustering structure for mining underlying cross-view discriminative information.
We also propose a novel unsupervised asymmetric metric learning model, i.e. CAMEL, for the metric initialization in DECAMEL.
CAMEL allows to explicitly model the view-specific conditions in unsupervised Re-ID \cite{2017_ICCV_asymmetric},
and thus the useful information of the cross-view person appearance variations can be exploited for the feature learning in DECAMEL.
}

\re{
\emph{\textbf{(3)}}
For large-scale view-extendable scenarios, we develop a method named View Clustering (VC) for better generalizability and scalability.
VC provides a flexible control on the generalizability versus ability to precisely model the view-specific conditions.
To evaluate our framework, we conduct extensive experiments on seven size-varying benchmark datasets.
Experimental results show that our model outperforms the state-of-the-art with noticeable margins,
indicating that the asymmetric modelling is effective in unsupervised Re-ID.
}

\section{Related Work}
\subsection{Unsupervised Re-ID Models}
While there are a lot of great works in supervised Re-ID [3-40],
unsupervised Re-ID still remains under-studied.
Existing attempts in unsupervised Re-ID can be classified into two categories: feature representation learning \cite{SDALF,CPS,SDC,GTS,ISR,RKSL} and
dictionary learning \cite{Dic,UDML,GL}.

\vspace{0.1cm}

\noindent
\textbf{Feature representation learning.}
This category of models mainly focuses on designing or learning discriminative and invariant features.
Farenzena et.al. \cite{SDALF} proposed to extract features containing three complementary parts of human appearance,
including color, spatial arrangement of colors and texture patches.
Cheng et.al. \cite{CPS} proposed to exploit the part-to-part correspondence.
They evaluate
the modified HSV characterization for each part and found the maximally stable color regions.
Zhao et.al. \cite{SDC} proposed to exploit salience information by
building dense correspondence and unsupervised salience learning.
Wang et.al. \cite{GTS} proposed to localize person foreground saliency and
remove busy background clutters.
Lisanti et.al. \cite{ISR} proposed to extract the weighted histogram of overlapping stripes (WHOS) feature,
and then applied the Iterative Re-Weighted Sparse Ranking (ISR) algorithm to generate the ranked list of gallery individuals.
Wang et.al. \cite{RKSL} proposed a CCA-based model to learn a feature subspace where within-view similar persons are
distant from each other and cross-view similar persons are close to each other.
Recently, Fan et.al. \cite{fan2017unsupervised} attempted to learn feature representation with a deep clustering network.

\vspace{0.1cm}

\noindent
\textbf{Dictionary learning.}
Dictionary learning aims to learn a dictionary with its atoms corresponding to some semantic elements.
The learned dictionary will be used to produce the new features which minimize the reconstruction error.
Kodirov et.al. \cite{Dic} proposed to formulate unsupervised Re-ID as a sparse dictionary
learning problem. To regularize the learned dictionary,
they learned it with graph Laplacian regularization, and iteratively updated the graph Laplacian matrix.
Then they took a further step \cite{GL}.
They propose to introduce an $l_1$-norm graph Laplacian to
jointly learn the graph and the dictionary, resulting in alleviation of the effects of data outliers \cite{GL}.
Peng et.al. \cite{UDML} proposed to learn a dictionary by unsupervised multi-task learning.
Different from \cite{Dic} and \cite{GL},
the learned dictionary consists of task-shared, task-specific and unique components.

Our model is very different from them in that:
\emph{(1)} Ours explicitly addresses the view-specific bias by
learning an asymmetric metric, i.e., learning view-specific transformation for each camera view.
Note that the RKSL \cite{RKSL} is based on Canonical Correlation Analysis (CCA) and also learns different projections for different views.
However, it does not address the view-specific bias problem,
because the CCA-based model RKSL learns two projections separately and inconsistently.
RKSL focuses on the correlations of cross-view samples by maximizing the correlation coefficients.
In the context of unsupervised learning (given no cross-view labelled pairs), not like ours,
RKSL is not able to discover the cross-view cluster structure and thus not suitable to mine the potential discriminative information across views.
In addition, no inherent consistency between two specific feature projections is considered,
while our model preserves cross-view consistency by learning an asymmetric metric under a cross-view consistency regularization.
\emph{(2)} Ours can jointly learn the feature representation and the asymmetric metric end to end.

\re{
This work is based on our preliminary work which presented CAMEL \cite{2017_ICCV_asymmetric}.
In addition to giving a more detailed description and analysis of the proposed linear model CAMEL,
the major differences are as follow:
\emph{(1)} We present a novel unsupervised deep framework DECAMEL.
Compared to the linear metric learning model CAMEL,
we propose the novel loss function, architecture and learning algorithm,
which together enable DECAMEL to perform joint learning of feature representation and metric in an unsupervised way.
By this means, compared to CAMEL, DECAMEL can learn a more compact cross-view cluster structure,
potentially facilitating mining the underlying cross-view discriminative information.
\emph{(2)} We propose the view clustering that can significantly improve the generalizability and scalability of DECAMEL.
\emph{(3)} We present more in-depth discussion and analysis on the proposed framework,
including a series of visual results which show intuitively and progressively how DECAMEL works.
Moreover, we conduct more extensive evaluations for comparisons and analysis.
}

\begin{figure*}[!t]
\centering
\includegraphics[width=0.7\linewidth]{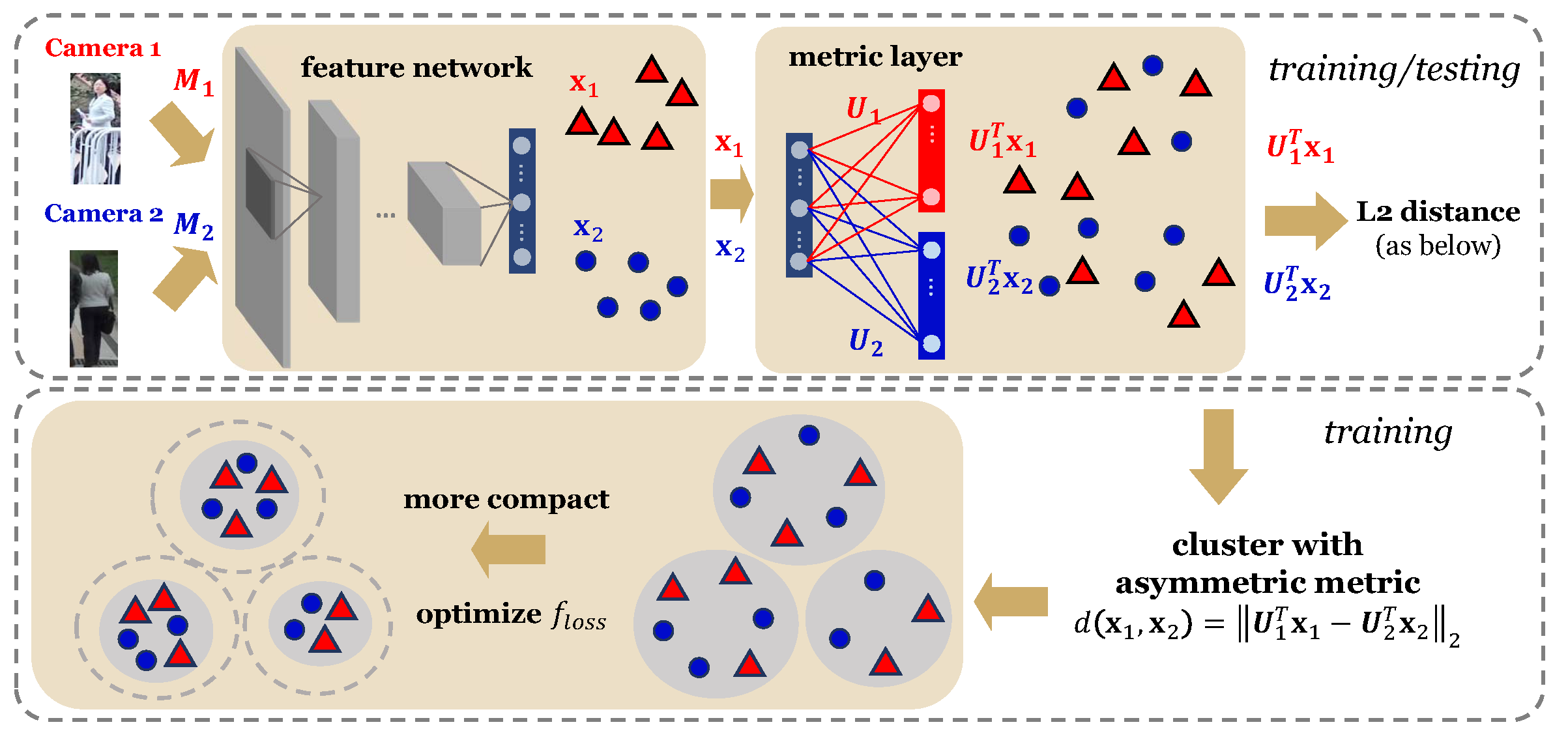}
\vspace{-0.4cm}
\caption{Illustration of our framework DECAMEL.
We follow the brown arrows to inspect it.
We extract features for person images by a deep network.
Due to view-specific conditions, the initial feature space has severe view-specific bias:
the red triangles (data points from Camera $1$) and blue circles (from Camera $2$) are far apart.
We perform CAMEL to learn an initial asymmetric metric.
In the shared space produced by the asymmetric metric, the view-specific bias is alleviated.
\re{By optimizing the proposed unsupervised loss, DECAMEL jointly learns the feature representation and asymmetric metric in an end-to-end manner.
During testing, pairwise distance can be computed by Eq. (\ref{equation:asymmetric}).}
(Best viewed in color).
}
\vspace{-0.2cm}
\label{figure:CAMEL}
\end{figure*}

\subsection{Unsupervised Metric Learning}
Although unsupervised metric learning has not been exploited in Re-ID,
there are a few works that exploit unsupervised metric learning in other fields \cite{2007_CVPR_AML,2015_NC_uNCA,um1,um2}.
Ye et.al. \cite{2007_CVPR_AML} proposed to jointly learn a metric and a clustering separation for better clustering results.
Qin et.al. \cite{2015_NC_uNCA} proposed to learn a metric based on regularized neighborhood component analysis for clustering analysis.
Cinbis et.al. \cite{um1} proposed to mine video information to find positive and negative pairs of faces,
and thus learning a metric for video face recognition.
Jiang et.al. \cite{um2} proposed a diffusion-based approach to improve an input similarity metric for vision tasks like
image segmentation and clustering.

Specifically, our model is closely related to the Adaptive Metric Learning (AML) \cite{2007_CVPR_AML}.
AML learns a transformation $\bm{G}$ to project the training data $\mathbf{x}$
onto a low dimensional space: $\hat{\mathbf{x}} = \bm{G}^{\mathrm{T}}\mathbf{x}$.
It minimizes the sum of squared error (SSE) in the subspace:
\begin{equation}
\small
SSE = \sum_{\hat{\mathbf{x}}} d_M(\hat{\mathbf{x}}, \hat{\mathbf{c}})^2,
\end{equation}
where $d_M$ is the Mahalanobis distance, and $\hat{\mathbf{c}}$ denotes the cluster centroid
to which $\hat{\mathbf{x}}$ belongs.

Our model is different from AML in that
ours performs cross-view clustering and explicitly models the view-specific bias in the context of unsupervised Re-ID.
We propose to alleviate the bias
by learning an asymmetric metric.
Furthermore, we propose to embed the metric learning into the deep neural networks for a joint learning
of feature representation and the asymmetric metric.

\re{The asymmetric modelling is also used in cross-modal retrieval \cite{cross-modal1, cca1, cca2},
where Canonical Correlation Analysis (CCA) \cite{cca} and Partial Least Squares (PLS) \cite{pls} based methods are arguably the most popular.
CCA and PLS also learn different projections for different modalities to induce a latent space,
where the correlation/covariance of pairwise cross-modal training samples is maximized.
However, they do not address the view-specific bias problem in person Re-ID since the projections are learned separately and inconsistently. Moreover, they require sufficient pairwise labelled data is required by these methods.}
There are also some works on supervised deep metric learning \cite{dml1, dml2, dml3, dml4, dml5, dml6}.
However, these models require substantial labelled data for training.
In contrast, our model can directly learn from unlabelled data
and thus it is free from requiring large amount of expensive labelled data.

\subsection{Unsupervised Deep Clustering Embedding}
Our model is also related to the unsupervised deep clustering embedding technique \cite{uc1,uc3,uc5}.
Xie et.al. \cite{uc1} proposed to jointly learn cluster membership and deep representation by minimizing the KL divergence
between original data distribution and a target distribution.
Yang et.al. \cite{uc3} proposed to jointly learn image clusters and deep representation using a convolutional neural network.
Wang et.al. \cite{uc5} proposed to learn a task-specific deep architecture based on sparse coding for clustering.

DECAMEL is different from all these models in that
it joins unsupervised asymmetric metric learning with cross-view clustering,
which is specifically designed for cross-view matching in Re-ID,
while others use standard symmetric metrics in clustering.

\vspace{-0.1cm}
\section{Approach}\label{section:approach}

In this section, we progressively develop our unsupervised Re-ID framework,
i.e., the \textbf{DE}ep \textbf{C}lustering-based \textbf{A}symmetric \textbf{ME}tric \textbf{L}earning (\emph{DECAMEL}).
Specifically,
our framework first learns an initial asymmetric metric by a linear unsupervised model (CAMEL) and then embeds the metric
into a deep network by jointly learning feature and metric.
These two steps are based on the asymmetric metric clustering,
so we first introduce the asymmetric metric clustering and develop the linear model.
Then, we introduce a novel loss function for the unsupervised deep joint learning.
We show an overview of our framework in Figure \ref{figure:CAMEL}.

\vspace{-0.2cm}
\subsection{Unsupervised Asymmetric Metric Learning}\label{section:approach_metricLearning}

Now let us dive into the details and develop the model step by step.
Under a general unsupervised Re-ID setting, we have $V$ camera views. From each of them, we have collected $N_v\;(v = 1,\cdots,V)$ images,
and thus we have $N = \sum_{v=1}^V N_v$ unlabelled images for training.
Here we assume that feature representation for the training images are given,
and the training set is denoted as $\mathcal{X} = \{\mathbf{x}_i, v_i\}_{i=1}^N$, where
$\mathbf{x}_i$ is the feature vector and $v_i$ denotes which camera view it comes from.
\re{Note that in visual surveillance, the camera view ``label'' $v_i$ is naturally available for each image,
since its straightforward to know by which camera an image is captured in a camera network.
Here we follow a popular assumption that during training and testing,
the camera views are the same [3, 5, 6, 8, 9, 11, 14, 15, 18-40, 44, 47-50].
We will also discuss a novel view-extendable setting in Sec. \ref{section:strategies}}


We are looking for some transformations to map
the data into a shared space, where we can better separate the images of different persons.
A natural idea is to decrease the intra-class (here, a class is a person)
discrepancy and enlarge the inter-class discrepancy.
In an unsupervised scenario,
however, we have no labelled data to strictly separate visually similar persons.
Therefore, we relax the original idea: we focus on gathering
similar person images together, and thereby separating relatively
dissimilar ones.
We formulate it by an objective like that of $k$-means clustering \cite{KMEANS}:
\begin{equation}\label{Eq0}
\small
\mathop{\min}_{\bm{U}, \{\mathbf{c}_k\}_{k=1}^K}f_{intra}=\frac{1}{N} \sum_{k=1}^K \sum_{i \in {\mathcal{C}_k}} \lVert \bm{U}^{\mathrm{T}}\mathbf{x}_i - \mathbf{c}_k \rVert^2,
\end{equation}
where $\bm{U}$ is the view-generic universal feature transformation, $K$ is the number of clusters,
$\mathbf{c}_k$ denotes the centroid of the $k$-th cluster and
$\mathcal{C}_k = \{ i | \bm{U}^{\mathrm{T}}\mathbf{x}_i \in k$-th cluster$\}$.

However, clustering can be largely affected
by the view-specific bias when applied in the cross-view problems.
In Re-ID, the variances across camera views like
different lighting conditions, human pose variations and occlusions \cite{2015_TCSVT_ASM} can be very dramatic.
They are disturbing or even dominating in searching the similar person images across views during
clustering procedure.
To address this problem,
we learn specific transformations for each view rather than a generic one,
to explicitly take the view-specific feature distortion into account and to alleviate the view-specific bias.
As discussed in Eq. (\ref{equation:asymmetric}), the idea can be further formulated by
\begin{equation}\label{Eq1}
\small
\begin{aligned}
\mathop{\min}_{\{\bm{U}_v\}_{v = 1}^V, \{\mathbf{c}_k\}_{k=1}^K}f_{intra}= &\frac{1}{N}\sum_{k=1}^K \sum_{i \in {\mathcal{C}_k}} \lVert \bm{U}^\mathrm{T}_{v_i}\mathbf{x}_i - \mathbf{c}_k \rVert^2\\
s.t.\qquad \bm{U}^\mathrm{T}_v\bm{\Sigma}_v\bm{U}_v &= \bm{I} \quad (v = 1,\cdots,V),
\end{aligned}
\end{equation}
where $\bm{U}_{v_i}$ is the specific feature transformation for the $v_i$-th camera view,
$v_i$ denotes which camera view $\mathbf{x}_i$ comes from,
$\bm{\Sigma}_v = \sum_{\mathbf{x}_t: v_t = v} \mathbf{x}_t\mathbf{x}_t^{\mathrm{T}} / N_v + \bm{I}$ is a covariance matrix,
and $\bm{I}$ represents the identity matrix which is used to avoid singularity of the covariance matrix.
The transformation for each instance $\mathbf{x}_i$ is determined by $v_i$.
The quasi-orthogonal constraints on $\bm{U}_v$ ensure that the model will
not simply give zero matrices. By jointly learning the asymmetric metric and cross-view clustering,
we actually realize an asymmetric metric clustering on Re-ID data across camera views.

Mathematically, if we minimize this objective function, every $\bm{U}_v$ will largely depend on the data distribution
from the $v$-th view. Since there is view-specific bias on each view, any pair of transformations---$\bm{U}_v$ and $\bm{U}_w$---could be arbitrarily different
according to the biases.
However, large inconsistencies among the learned transformations are not
what we expect, since these transformations are with respect to person images from different views.
Although under different conditions, the subjects are human beings,
and thus they are inherently correlated and not heterogeneous.
Therefore, largely different projection basis pairs would fail to capture the
discriminative nature of the person images.

Hence, to strike a balance between the ability to preserve the cross-view consistency and
the ability to alleviate view-specific bias, we add a cross-view consistency regularization term
to our objective function.
The cross-view consistency regularization penalizes the discrepancy between any pair of correlated transformation basis $\mathbf{u}_v^c$ and $\mathbf{u}_w^c$,
where $\mathbf{u}_v^c$ is the $c$-th column of $\bm{U}_v$.
Thus, we formulate it as:
\begin{equation}
\small
f_{consistency} = \sum_{v, w}\sum_c \lVert \mathbf{u}_v^c - \mathbf{u}_w^c \rVert_2^2 = \sum_{v, w} \lVert \bm{U}_v - \bm{U}_w \rVert_F^2,
\end{equation}
where $\lVert\cdot\rVert$ is the Frobenius norm of a matrix.
And then, our optimization task is given by
\begin{equation}\label{equation:f_obj1}
\small
\begin{aligned}
\mathop{\min}_{\{\bm{U}_v\}_{v = 1}^V, \{\mathbf{c}_k\}_{k=1}^K}
f_{obj} &= f_{intra} + \lambda f_{consistency} \\
= \frac{1}{N}\sum_{k=1}^K \sum_{i \in {\mathcal{C}_k}} &\lVert \bm{U}^\mathrm{T}_{v_i}\mathbf{x}_i - \mathbf{c}_k \rVert^2 + \lambda\sum_{v, w} \lVert \bm{U}_v - \bm{U}_w \rVert_F^2 \\
s.t. \qquad \bm{U}_v^{\mathrm{T}}\bm{\Sigma}_v&\bm{U}_v = \bm{I} \quad (v = 1,\cdots,V),
\end{aligned}
\end{equation}
where $\lambda$ is the cross-view regularizer.
We call the above model the \textbf{C}lustering-based \textbf{A}symmetric \textbf{ME}tric \textbf{L}earning (\emph{CAMEL}).

We will show by an illustration that the asymmetric metric can alleviate the view-specific bias
in the Re-ID problem in Sec. \ref{section:asymmetricAligns},
and show that the cross-view consistency regularization contributes much in our framework in Sec. \ref{section:parameterDiscussion}.
\begin{center}
\begin{tabular}{|p{0.92\linewidth}|}\hline
\rule{0pt}{3ex}
\noindent \textbf{Remark 1: Cross-view Consistency Regularization for the Metric}.
We note that although asymmetric metric learning has been successfully applied in supervised Re-ID \cite{f2, 2015_TCSVT_ASM},
it is a pseudo metric rather than a strict metric \cite{2015_TCSVT_ASM},
because it may not meet the coincidence property: given two identical feature vectors $\mathbf{x}_i$ and $\mathbf{x}_j$ ($\mathbf{x}_i = \mathbf{x}_j$) from different camera views $v_i$ and $v_j$,
the asymmetric metric may not guarantee $d(\mathbf{x}_i, \mathbf{x}_j) = 0$.
In this aspect, the cross-view consistency regularization plays \re{a} role to control
an upper bound of coincidence discrepancy.
In fact, according to the Cauchy Inequality, we have
\begin{equation}
\small
\begin{aligned}
d(\mathbf{x}_i, \mathbf{x}_j) &= \lVert \bm{U}_{v_i}^\mathrm{T}\mathbf{x}_i - \bm{U}_{v_j}^\mathrm{T}\mathbf{x}_j \rVert_2
= \lVert \bm{U}_{v_i}^\mathrm{T}\mathbf{x}_i - \bm{U}_{v_j}^\mathrm{T}\mathbf{x}_i \rVert_2 \\
&\leq \lVert \mathbf{x}_i \rVert_2 \cdot \lVert \bm{U}_{v_i} - \bm{U}_{v_j} \rVert_F.
\end{aligned}
\end{equation}
The cross-view consistency regularization controls $\lVert \bm{U}_{v_i} - \bm{U}_{v_j} \rVert_F$ which is a scaled upper bound of the coincidence discrepancy.
Thus, it makes the learned asymmetric metric more mathematically principled and rigorous.
\\\hline
\end{tabular}
\end{center}

\subsection{Deep Unsupervised Asymmetric Metric Embedding}
Based on CAMEL and the foregoing analysis,
we can further propose our framework, named the \textbf{DE}ep \textbf{C}lustering-based \textbf{A}symmetric \textbf{ME}tric \textbf{L}earning (\emph{DECAMEL}).
DECAMEL embeds the unsupervised asymmetric metric into the network by jointly learning the feature representation and the asymmetric metric.

In the last subsection we assume that we have extracted features for the training images.
Here, we specify the feature extractor function $f$
as a deep convolutional network, which is parameterized by $\mathbf{\Theta}$.
So, the feature representation is given by $\mathbf{x}_i = f(\mathbf{M}_i;\mathbf{\Theta})$,
where $\mathbf{M}$ is an image from our training image set $\mathcal{M} = \{\mathbf{M}_i, v_i\}_{i=1}^N$.
Our goal is to develop an end-to-end framework, $g(\mathbf{M}_i, v_i;\mathbf{\Theta}, \{\bm{U}_v\}_{v = 1}^V)$,
which includes the feature extraction and the unsupervised asymmetric metric learning.

Recall the objective in Eq. (\ref{equation:f_obj1}),
where in CAMEL we aim to minimize it under \re{constraints}.
To avoid trivial implementation and better adapt to the back-propagating algorithm in the optimization procedure,
we develop a soft version of \re{constraints},
by replacing the \re{constraints} with a regularization term $f_{\re{constraint}}$ similar to $f_{consistency}$:
\begin{equation}
\small
f_{constraint} = \sum_{v=1}^V\lVert\bm{U}_v^{\mathrm{T}}\bm{\Sigma}_v\bm{U}_v - \bm{I}\rVert_F^2.
\end{equation}
This technology is widely used in the machine learning problems.
For example, the well-known Tikhonov regularization is a soft version
of the Ivanov regularization \cite{vapnik1998statistical}.
By this way, we have our loss function of DECAMEL as follow:
\begin{equation}\label{equation:loss}
\small
\begin{aligned}
f_{loss} &= f_{intra} + \lambda f_{consistency} + \gamma f_{constraint} \\
&= \frac{1}{N}\sum_{k=1}^K \sum_{i \in {\mathcal{C}_k}} \lVert \bm{U}^\mathrm{T}_{v_i}\mathbf{x}_i - \mathbf{c}_k \rVert^2 + \lambda\sum_{u, w} \lVert \bm{U}_v - \bm{U}_w \rVert_F^2 \\
&+ \gamma \sum_{v=1}^V\lVert\bm{U}_v^{\mathrm{T}}\bm{\Sigma}_v\bm{U}_v - \bm{I}\rVert_F^2,
\end{aligned}
\end{equation}
where $\gamma$ is the soft-constraint parameter. In our experiments we set $\gamma = 10$ to basically
ensure the constraints. \re{We empirically find this hyper-parameter needs no exhaustive tuning across datasets.}

CAMEL is a linear metric learning model.
Intrinsically, it cannot discover the underlying non-linear cross-view cluster structure of the feature representation.
In addition, the feature extraction is separated and independent from metric in CAMEL.
This leads to sub-optimality
because the feature representation might have the capacity to be further improved according to the metric.
In contrast, DECAMEL addresses these problems naturally.
It jointly learns the feature representation and the asymmetric metric
with non-linearity capacity.
Thus, it learns a better cross-view cluster structure.


\begin{figure*}[!t]
\centering
\subfigure[Feature representation distribution\label{figure:originalDistribution}]{
\includegraphics[width=0.25\linewidth, height=0.25\linewidth]{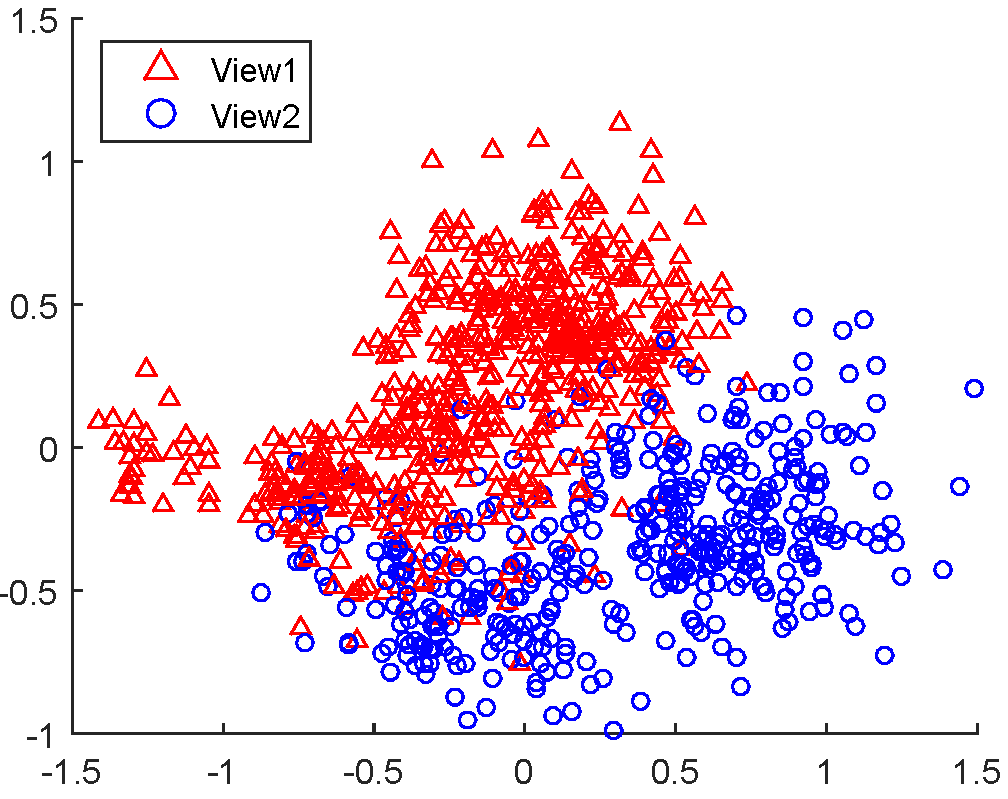}
}
\subfigure[Using symmetric metric\label{figure:symmetricDistribution}]{
\includegraphics[width=0.25\linewidth, height=0.25\linewidth]{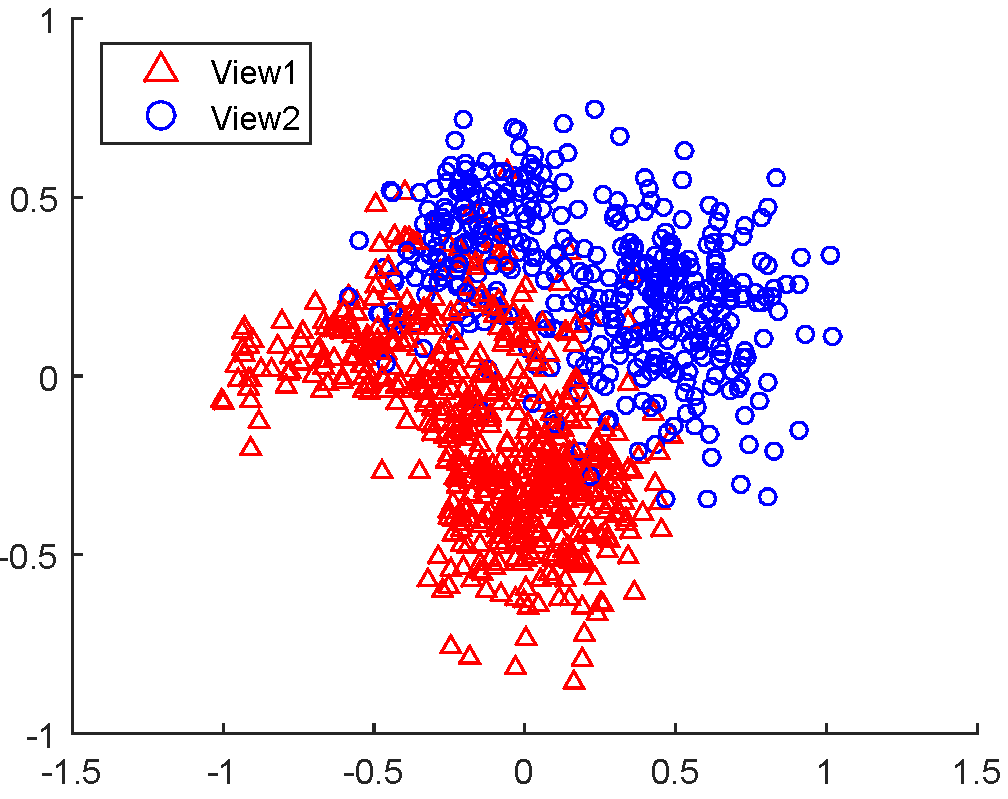}
}
\subfigure[Using asymmetric metric\label{figure:asymmetricDistribution}]{
\includegraphics[width=0.25\linewidth, height=0.25\linewidth]{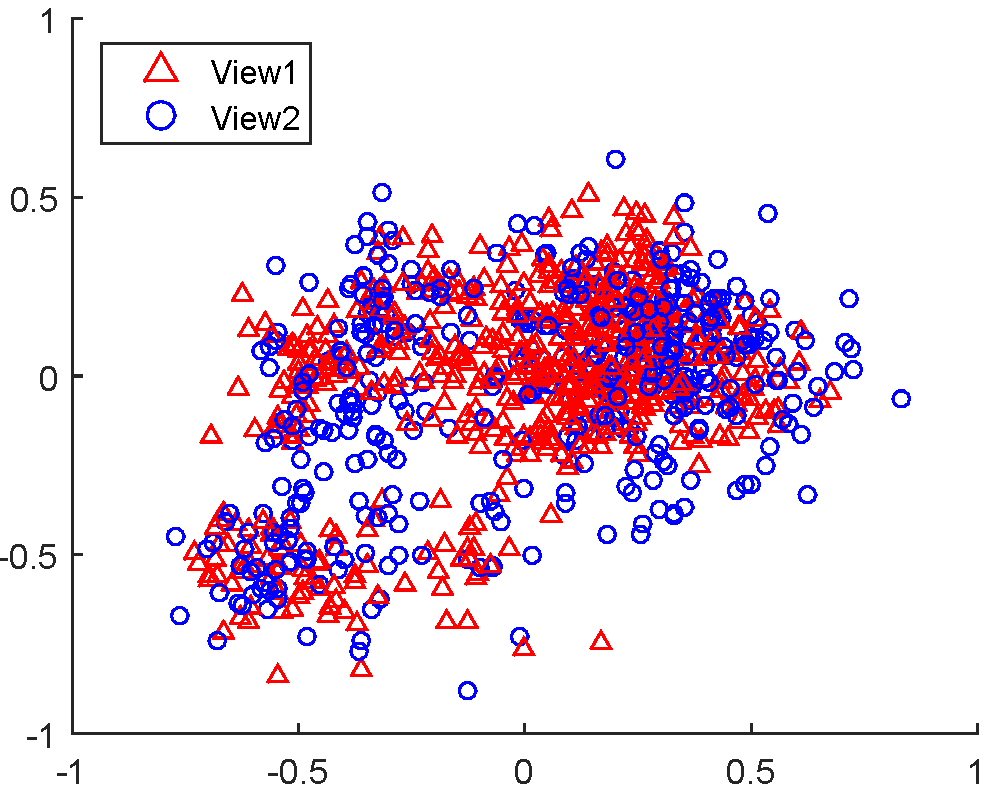}
}
\vspace{-0.2cm}
\caption{Illustration of asymmetric metric alleviating view-specific bias.
The data is randomly sampled from the SYSU dataset \cite{2015_TCSVT_ASM}.
We performed PCA for visualization.
Blue circles and red triangles represent data points from two camera views.
\textbf{\emph{(a)}} Cross-view data distribution in original feature representation space.
View-specific bias is severe here, since one can easily draw a boundary to roughly separate the circles and triangles.
\textbf{\emph{(b)}} Distribution in the shared space after projected by the learned view-generic transformation (symmetric metric).
The bias is not alleviated.
\textbf{\emph{(c)}} Distribution in the shared space after projected by the learned view-specific transformations (asymmetric metric).
The bias is largely alleviated.
(Best viewed in color).
}
\label{figure:distribution}
\end{figure*}

\begin{figure*}[!t]
\centering

\includegraphics[width=0.85\linewidth]{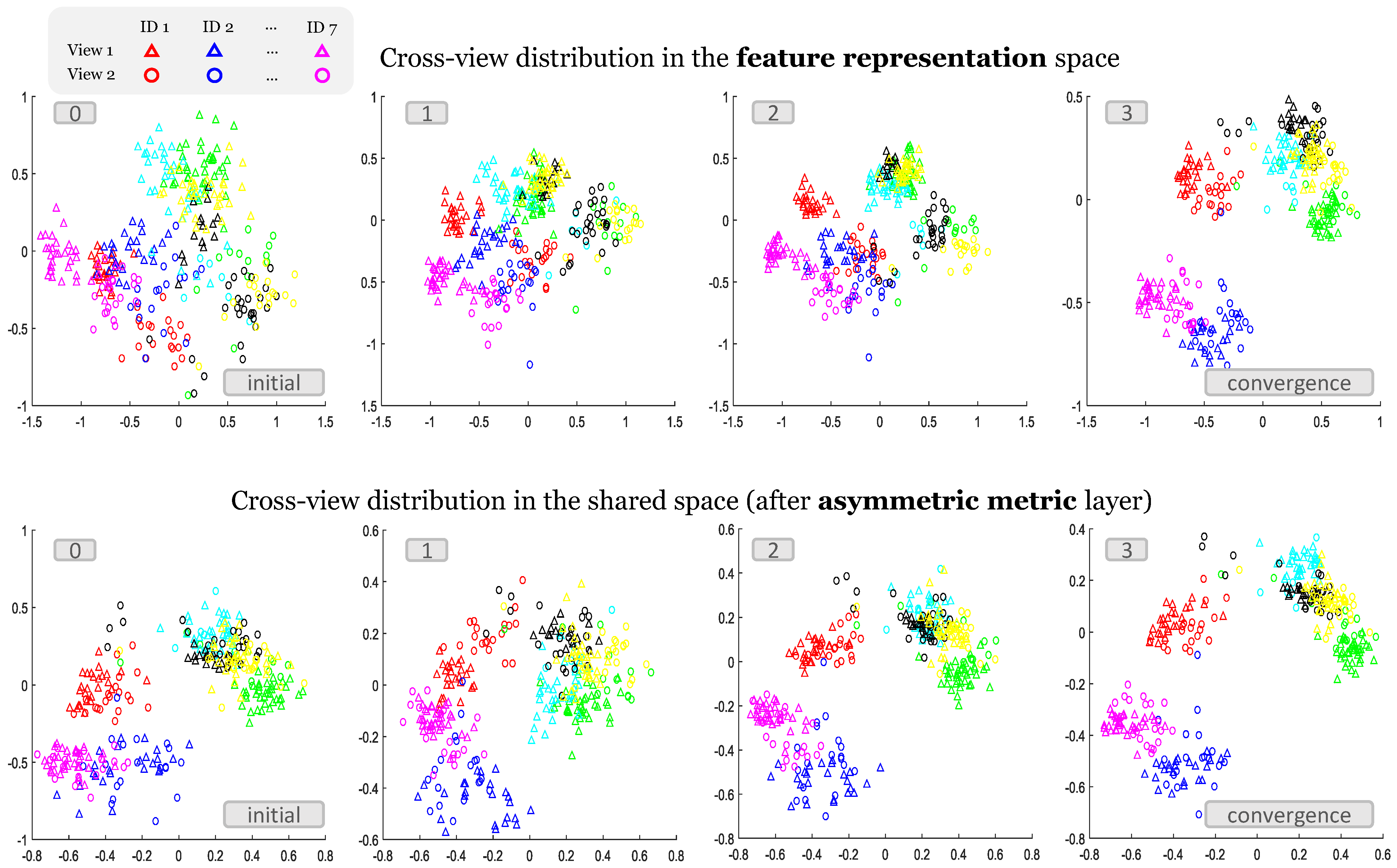}
\vspace{-0.4cm}
\caption{Illustration of DECAMEL learning the better cross-view cluster structure via jointly learning the feature representation and the asymmetric metric.
\re{We perform PCA for visualization.
Images of an identity are indicated by a specific color (e.g. all red triangles and circles are images of the first identity in the feature space).}
The numbers in the upper left of each figure indicates different stages, from initial to convergence.
The two figures in each column are synchronous and corresponding to each other.
Data points are only a subset from those in Figure \ref{figure:distribution} for clarity.
Specifically, the initial stages (the leftmost column) are subsets of Figure \ref{figure:originalDistribution} and Figure \ref{figure:asymmetricDistribution},
respectively.
(Best viewed in color and please refer to the text in Sec. \ref{section:DECAMELEmbeds} for more analysis.
Please zoom in for better visual quality).}
\label{figure:FM}
\end{figure*}

\subsection{Optimization and Algorithm}
Before performing the gradient descent to train DECAMEL,
we first learn an unsupervised asymmetric metric $\{\bm{U}_v\}_{v = 1}^V$ by CAMEL to initialize the metric layer of DECAMEL,
as well as the cluster results $\{\mathbf{c}_k\}_{k=1}^K$.
So we first introduce how to optimize the objective of CAMEL.
We will see that the metric initialization by CAMEL contributes a lot to the whole framework in Sec \ref{section:metricInitialization}.

\subsubsection{Metric Initialization by CAMEL}

For convenience, let $\mathbf{y}_i = \bm{U}^{\mathrm{T}}_{v_i}\mathbf{x}_i$ and $\bm{Y}=[\mathbf{y}_1, \cdots, \mathbf{y}_N]$.
We rewrite our objective function using trace instead of sum.
The first term $f_{intra}$ can be rewritten as \cite{NMF}
\begin{equation}\label{equation:f_intra_trace1}
\begin{aligned}
\small
\sum_{k=1}^K \sum_{i \in {\mathcal{C}_k}} \lVert \bm{U}^\mathrm{T}_{v_i}\mathbf{x}_i - \mathbf{c}_k \rVert^2
=[\mathrm{Tr}(\bm{Y}^{\mathrm{T}}\bm{Y})-\mathrm{Tr}(\bm{H}^{\mathrm{T}}\bm{Y}^{\mathrm{T}}\bm{YH})], \\
\end{aligned}
\end{equation}
where
\begin{equation}\label{equation:H}
\small
\bm{H} =
\begin{bmatrix}
\mathbf{h}_1,...,\mathbf{h}_K
\end{bmatrix}
,\quad \mathbf{h}_k^{\mathrm{T}}\mathbf{h}_l =
\begin{cases}
0 & k\neq l \\
1 & k= l
\end{cases},
\end{equation}
and
\begin{equation}\label{equation:h_k}
\small
\mathbf{h}_k =
\begin{bmatrix}
0,\cdots,0,1,\cdots,1,0,\cdots,0,1,\cdots
\end{bmatrix}
^{\mathrm{T}}/\sqrt{n_k}
\end{equation}
is an indicator vector with the $i$-th entry corresponding to the instance $\mathbf{y}_i$,
indicating that $\mathbf{y}_i$ is in the $k$-th cluster if the corresponding entry \re{is non-zero}.

Now we construct an assignment function $\xi:\mathcal{X}\rightarrow\mathbb{R}^{Vd}$
where $d$ denotes the feature dimension:
\begin{equation}
\small
\xi(\{\mathbf{x}, v\}) =
\begin{bmatrix}
\mathbf{0}^{\mathrm{T}}, \cdots, \mathbf{0}^{\mathrm{T}}, \mathbf{x}^{\mathrm{T}}, \mathbf{0}^{\mathrm{T}}, \cdots, \mathbf{0}^{\mathrm{T}}
\end{bmatrix}
^{\mathrm{T}},
\end{equation}
where $\mathbf{0}$ is a zero column vector which has the same size as $\mathbf{x}$,
and $\mathbf{x}$ is placed in the $v$-th ``entry''. Then we can construct
$\widetilde{\bm{X}} = [\widetilde{\mathbf{x}}_1, \cdots, \widetilde{\mathbf{x}}_N] \in \mathbb{R}^{Vd \times N}$,
where $\widetilde{\mathbf{x}}_i = \xi(\{\mathbf{x}_i, v_i\})$.
Besides, we construct
\begin{equation}
\small
\widetilde {\bm{U}} =
\begin{bmatrix}
\bm{U}_1^{\mathrm{T}}, \cdots, \bm{U}_V^{\mathrm{T}}
\end{bmatrix}
^{\mathrm{T}}
,
\end{equation}
so that
\begin{equation}\label{equation:Y}
\bm{Y} = \widetilde{\bm{U}}^{\mathrm{T}}\widetilde{\bm{X}}.
\end{equation}
Substitute Eq. (\ref{equation:Y}) into Eq. (\ref{equation:f_intra_trace1}) and thus $f_{intra}$ becomes
\begin{equation}
\begin{aligned}
\small
f_{intra}
=&\frac{1}{N}\mathrm{Tr}(\widetilde {\bm{X}}^{\mathrm{T}}\widetilde {\bm{U}}\widetilde {\bm{U}}^{\mathrm{T}}\widetilde {\bm{X}})
-\frac{1}{N}\mathrm{Tr}({\bm{H}}^{\mathrm{T}}\widetilde {\bm{X}}^{\mathrm{T}}\widetilde {\bm{U}}\widetilde {\bm{U}}^{\mathrm{T}}\widetilde {\bm{X}}\bm{H}).
\end{aligned}
\end{equation}
As for the second term, we can also rewrite $f_{consistency}$ as
\begin{equation}
f_{consistency} = \sum_{v, w} \lVert \bm{U}_v-\bm{U}_w\rVert_F^2 = \mathrm{Tr}(\widetilde{\bm{U}}^{\mathrm{T}}\bm{D\widetilde U}),
\end{equation}
where
\begin{equation}
\small
\bm{D} =
\begin{bmatrix}
(V-1)\bm{I}& -\bm{I}& -\bm{I}&\cdots &-\bm{I} \\
-\bm{I}& (V-1)\bm{I}& -\bm{I}&\cdots &-\bm{I} \\
\vdots&\vdots&\vdots&\vdots&\vdots \\
-\bm{I}&  -\bm{I}& -\bm{I}& \cdots&(V-1)\bm{I}
\end{bmatrix}\in \re{\mathbb{R}^{Vd\times Vd}}
\end{equation}
\re{has $V\times V$ block entries}
Then, it is reasonable to relax the constraints
\begin{equation}
\small
\bm{U}_v^{\mathrm{T}}\bm{\Sigma}_v\bm{U}_v = \bm{I} \quad (v = 1,\cdots,V)
\end{equation}
to
\begin{equation}
\small
\sum_{v=1}^V \bm{U}_v^{\mathrm{T}}\bm{\Sigma}_v\bm{U}_v = \widetilde {\bm{U}}^{\mathrm{T}}\widetilde{\bm{\Sigma}}\widetilde {\bm{U}} = V\bm{I},
\end{equation}
where $\widetilde{\bm{\Sigma}} = diag(\bm{\Sigma}_1, \cdots, \bm{\Sigma}_V)$,
because what we expect is to prevent each $\bm{U}_v$ from shrinking to a zero matrix.
The relaxed version of constraints is able to satisfy such a need, and it
allows more elegant optimization.
By now we can rewrite our optimization task as \re{follows}:
\begin{equation}\label{equation:f_obj2}
\small
\begin{aligned}
\mathop{\min}_{\widetilde{\bm{U}}, \bm{H}}f_{obj}
&= \frac{1}{N}\mathrm{Tr}(\widetilde {\bm{X}}^{\mathrm{T}}\widetilde {\bm{U}}\widetilde {\bm{U}}^{\mathrm{T}}\widetilde {\bm{X}})
+\lambda\mathrm{Tr}(\widetilde{\bm{U}}^{\mathrm{T}}\bm{D\widetilde U}) \\
 &- \frac{1}{N}\mathrm{Tr}({\bm{H}}^{\mathrm{T}}\widetilde {\bm{X}}^{\mathrm{T}}\widetilde {\bm{U}}\widetilde {\bm{U}}^{\mathrm{T}}\widetilde {\bm{X}}\bm{H})
 \\
&s.t.\qquad \widetilde {\bm{U}}^{\mathrm{T}}\widetilde{\bm{\Sigma}}\widetilde {\bm{U}} = V\bm{I}.
\end{aligned}
\end{equation}
We can easily find that our objective function
is non-convex. Fortunately, in the form of Eq. (\ref{equation:f_obj2}),
we can find that once $\bm{H}$ is fixed, Lagrange's method can be applied to
our optimization task. And from Eq. (\ref{equation:f_intra_trace1}),
we can find that it is exactly the objective of $k$-means clustering \cite{KMEANS} with respect to $\mathbf{y}_i$ once $\widetilde{\bm{U}}$ is fixed.
Thus, we can adopt an EM-like alternating algorithm to solve the optimization problem.

\vspace{0.15cm}
\noindent \textbf{Fix $\bm{H}$ and optimize $\widetilde{\bm{U}}$}.
After fixing $\bm{H}$ and applying the method
of Lagrange multiplier, our optimization task (\ref{equation:f_obj2})
is transformed into an eigen-decomposition problem as follow:
\begin{equation}\label{equation:eigen1}
\small
\bm{G}\mathbf{u} = \eta \mathbf{u},
\end{equation}
where $\eta$ is the Lagrange multiplier (and also is the eigenvalue here) and
\begin{equation}\label{equation:eigen2}
\small
\bm{G} = \widetilde{\bm{\Sigma}}^{-1}(\lambda \bm{D}+\frac{1}{N}\widetilde{\bm{X}}\widetilde{\bm{X}}^{\mathrm{T}}-\frac{1}{N}\widetilde{\bm{X}}\bm{HH}^{\mathrm{T}}\widetilde{\bm{X}}^{\mathrm{T}}).
\end{equation}
$\widetilde{\bm{U}}$ can be obtained by solving this eigen-decomposition problem.

\vspace{0.15cm}
\noindent \textbf{Fix $\widetilde{\bm{U}}$ and optimize $\bm{H}$}. As for the optimization of $\bm{H}$, we can simply fix $\widetilde{\bm{U}}$
and conduct $k$-means clustering in the learned space. Each column of $\bm{H}$,
$\mathbf{h}_k$, is thus constructed by Eq. (\ref{equation:h_k}) according to the $k$-means clustering result.

\subsubsection{Optimizing DECAMEL by Gradient \re{Descent}}

After obtaining the initial $\{\bm{U}_v\}_{v=1}^V$ and $\{\mathbf{c}_k\}_{k = 1}^K$,
we can optimize DECAMEL.
We adopt the stochastic gradient descent (SGD) to optimize DECAMEL.
The gradients are
\begin{equation}\label{equation:gradient1}
\begin{aligned}
\small
\frac{\partial{f_{loss}}}{\partial{\bm{U}_v^{\mathrm{T}}\mathbf{x}}} = 2(\bm{U}_v^{\mathrm{T}}\mathbf{x} - \mathbf{c}_k) \\
\end{aligned}
\end{equation}
and
\begin{equation}\label{equation:gradient2}
\small
\begin{aligned}
\frac{\partial{f_{loss}}}{\partial{\bm{U}_v}} &= 2(\mathbf{x}\mathbf{x}^{\mathrm{T}}\bm{U}_v - \mathbf{x}\mathbf{c}_k),
\end{aligned}
\end{equation}

\begin{equation}\label{equation:gradient3}
\small
\begin{aligned}
\frac{\partial{f_{loss}}}{\partial{\widetilde{\bm{U}}}} &= 2\lambda\bm{D}\widetilde{\bm{U}} + 4\gamma\widetilde{\bm{\Sigma}}\widetilde{\bm{U}}(\widetilde{\bm{U}}^{\mathrm{T}}\widetilde{\bm{\Sigma}}\widetilde{\bm{U}}-V\bm{I}).
\end{aligned}
\end{equation}

We note that the gradient in Eq. (\ref{equation:gradient1}) is with respect to each sample
and the gradients in Eq. (\ref{equation:gradient2}) and Eq. (\ref{equation:gradient3}) are with respect to the asymmetric metric.
Thus, we refer to Eq. (\ref{equation:gradient1}) as the sample gradient
and refer to Eq. (\ref{equation:gradient2}) and Eq. (\ref{equation:gradient3}) as the metric gradient in the following Remark 2.
We show the main algorithm of DECAMEL in Algorithm \ref{AlgDeCamel}.
\re{Note that $f_{obj}$ is guaranteed to converge as proved in \cite{2017_ICCV_asymmetric}.
It typically reaches convergence within $20$ iterations.}

\begin{table}[t]
\normalsize
\begin{tabular}{|p{0.92\linewidth}|}\hline
\rule{0pt}{3ex}
\noindent  \textbf{Remark 2: Explanation for Deep Metric Embedding}.
We can see from Eq. (\ref{equation:gradient1}), (\ref{equation:gradient2}) and (\ref{equation:gradient3}) that, the sample gradient flows over the whole network,
while the metric gradient only flows to the metric.
However,
the metric is actually embedded into the sample gradient:
according to the chain rule, the sample gradient for the feature extractor parameter $\mathbf{\Theta}$ is
\begin{equation}
\small
\begin{aligned}
\frac{\partial{f_{loss}}}{\partial{\mathbf{\Theta}}} &= \frac{\partial{f_{loss}}}{\partial{f(\mathbf{M}; \mathbf{\Theta})}}
\frac{\partial{f(\mathbf{M}; \mathbf{\Theta})}}{\partial{\mathbf{\Theta}}} =
\frac{\partial{f_{loss}}}{\partial{\mathbf{x}}}
\frac{\partial{f(\mathbf{M}; \mathbf{\Theta})}}{\partial{\mathbf{\Theta}}} \\
&= \frac{\partial{f_{loss}}}{\partial{\bm{U}_v^{\mathrm{T}}\mathbf{x}}}
\frac{\partial{\bm{U}_v^{\mathrm{T}}\mathbf{x}}}{\partial{\mathbf{x}}}
\frac{\partial{f(\mathbf{M}; \mathbf{\Theta})}}{\partial{\mathbf{\Theta}}} \\
&= 2(\bm{U}_v^{\mathrm{T}}\mathbf{x} - \mathbf{c}_k)\bm{U}_v^{\mathrm{T}} \cdot \frac{\partial{f(\mathbf{M}; \mathbf{\Theta})}}{\partial{\mathbf{\Theta}}}.
\end{aligned}
\end{equation}
Thus, the metric $\bm{U}_v^{\mathrm{T}}$ is back-propagated to the whole network.
As we will see in Sec. \ref{section:DECAMELEmbeds}, the jointly learned feature bears resemblance to the metric.
Furthermore, we will also see in Sec. \ref{section:component_wise} that this improves the cross-view discriminability of the feature.
These observations seem like the metric is being ``embedded'' into the feature,
and thus we refer to it as the ``deep metric embedding''.
\\\hline
\end{tabular}
\end{table}

\begin{algorithm}[t]\label{AlgDeCamel}
\scriptsize
\caption{DECAMEL}
\SetKwInOut{Input}{Input}
\SetKwInOut{Output}{Output}
\Input{The training images $\mathcal{M}$, the deep feature extractor $f(\cdot; \mathbf{\Theta})$}
\textbf{Training}: \\
\emph{Metric initialization}: \\
Extract feature representations using $f$ to obtain the initial feature set $\mathcal{X}$. \\
Conduct $k$-means clustering in $\mathcal{X}$ to obtain $\{\mathbf{c}_k\}_{k = 1}^K$
and to initialize $\bm{H}$ according to Eq. (\ref{equation:H}) and (\ref{equation:h_k}). \\
Fix $\bm{H}$ and solve the eigen-decomposition problem described by Eq. (\ref{equation:eigen1}) and (\ref{equation:eigen2})
to construct $\widetilde{\bm{U}}$. \\
$t \leftarrow 1$ where $t$ denotes each step in the following loop. \par
\While{$\{f_{obj}^t\}$ not converged}
{
\begin{itemize}[leftmargin=*]
\setlength{\topsep}{1ex}
\setlength{\itemsep}{-0.1ex}
\setlength{\parskip}{0.1\baselineskip}
\vspace{0.1cm}
\item Alternate fixing $\widetilde{\bm{U}}$ and $\bm{H}$ while optimizing the other.
\item $t\leftarrow t+1$.
\end{itemize}
}
Decompose $\widetilde{\bm{U}}$ to obtain $\{\bm{U}_v\}_{v=1}^V$.\\
Initialize the deep framework $g(\cdot, \cdot;\mathbf{\Theta}, \{\bm{U}_v\}_{v = 1}^V)$ using $\mathbf{\Theta}$ and $\{\bm{U}_v\}_{v = 1}^V$. \\
\emph{End-to-end joint learning}: \\
Update $\{\mathbf{c}_k\}_{k = 1}^K$ from $\bm{H}$ according to Eq. (\ref{equation:H}) and (\ref{equation:h_k}).\\
\While{maximum iteration not reached}
{
\begin{itemize}[leftmargin=*]
\setlength{\topsep}{1ex}
\setlength{\itemsep}{-0.1ex}
\setlength{\parskip}{0.1\baselineskip}
\vspace{0.1cm}
\item Update $\mathbf{\Theta}$ and $\{\bm{U}_v\}_{v = 1}^V$ by performing SGD using the gradients in \\ Eq. (\ref{equation:gradient1}), (\ref{equation:gradient2}) and (\ref{equation:gradient3}).
\item Update $\{\mathbf{c}_k\}_{k = 1}^K$ while fixing $\mathbf{\Theta}$ and $\{\bm{U}_v\}_{v = 1}^V$.
\end{itemize}
}
\re{\textbf{Testing}: \\
Given two testing images $\{\mathbf{M}_i, v_i\}$ and $\{\mathbf{M}_j, v_j\}$,
the distance/dissimilarity is computed by $||g(\mathbf{M}_i, v_i) - g(\mathbf{M}_j, v_j)||_2$.}
\end{algorithm}

\begin{algorithm}[t]\label{AlgDECAMELVC}
\scriptsize
\caption{DECAMEL with View Clustering}
\SetKwInOut{Input}{Input}
\Input{The training images $\mathcal{M}$, the deep feature extractor $f(\cdot; \mathbf{\Theta})$}
\textbf{Training}: \\
Compute the view representations $\{\mathbf{w}_v\}_{v=1}^V$ by Eq. (\ref{equation:viewRepr}) . \\
Conduct $k$-means clustering in $\{\mathbf{w}_v\}_{v=1}^V$ to obtain the cluster separation
$\mathcal{B}_j = \{ v | \mathbf{w}_v \in j$-th cluster$\}$. \\
For each image $M_i$ in the training set, reassign a view label $v_i^\prime \leftarrow j$ where $v_i \in \mathcal{B}_j$ to it.
So we now have $1 \leq v_i^\prime \leq J$ (the number of view clusters). \\
Feed $\mathcal{M}^\prime = \{\mathbf{M}_i, v_i^\prime\}$ to Algorithm \ref{AlgDeCamel}
to train a deep framework $g(\cdot, \cdot;\mathbf{\Theta}, \{\bm{U}_j\}_{j = 1}^J)$. \\
Use the learned feature extractor $g(\cdot; \mathbf{\Theta})$ to compute view prototypes/centroids $\{\mathbf{b}_j\}_{j=1}^J$. \\
\textbf{Testing for an unseen view $u$}: \\
Extract the view representation $\mathbf{w}_u$ using the learned feature extractor $g(\cdot; \mathbf{\Theta})$. \\
Assign this view to a view prototype $j$ where $j = \arg\min_{j} \lVert\mathbf{w}_u - \mathbf{b}_j\rVert_2$. \\
Assign a view label $j$ to all testing images from this view. \\
Follow the testing procedure in Algorithm \ref{AlgDeCamel}.

\end{algorithm}

\re{\subsection{View Clustering: Generalizing to unseen views}\label{section:strategies}}

\re{
In the beginning of this section we follow the conventional Re-ID setting that assumes the training and testing camera views are the same.
However, some realistic large-scale applications might need to be view-extendable,
i.e. new unseen camera views might be added to the surveillance network after training.
In this case, the \emph{generalizability} to new views becomes important.
We propose a general method toward view-extendable scenarios for achieving better generalizability.
The proposed method does not need to re-train the model when new views are added after training.
}

\re{
The main idea is that instead of learning feature transformations for each camera view,
we can learn transformations for some \emph{generalizable view prototypes},
which shall cover the most typical view-specific conditions.
Then, if a new, unseen view is added after training,
we can assign the new view to a view prototype and thus use the corresponding feature transformation.
We elaborate our strategy in the following.
}

\re{
To better motivate our method, we start from defining pairwise dissimilarity/distance $d_V(\cdot,\cdot)$ of two camera views.
As our model addresses view-specific bias in an overall \emph{view level},
a straightforward idea is to define $d_V(\cdot,\cdot)$ as the distance between \emph{distributions} of images from the two views.
We adopt the simplified 2-Wasserstein distance \cite{2017_ICML_WGAN, 2017_Arxiv_BEGAN}\footnote{This simplified 2-Wasserstein distance
makes a Gaussian assumption over the sample features, which is also empirically observed in our experiments.
We refer the reader to \cite{2017_Arxiv_BEGAN} for further justification.},
which has been shown effective and easy-to-compute in many vision tasks \cite{2017_Arxiv_BEGAN, 2018_TPAMI_WCNN}, defined as:
\begin{equation}\label{equation:distanceViews}
d_V(View_u, View_v)^2 = \frac{1}{2}(\lVert\mathbf{m}_u - \mathbf{m}_v\rVert^2_2+\lVert\boldsymbol{\sigma}_u - \boldsymbol{\sigma}_v\rVert^2_2),
\end{equation}
where $\mathbf{m}_v$ is the mean vector of all training sample features (extracted by the feature network) from $View_v$,
$\boldsymbol{\sigma}_v$ is the corresponding standard deviation vector,
and $\mathbf{m}_u, \boldsymbol{\sigma}_u$ are similarly defined.
From Eq. (\ref{equation:distanceViews}) we can define the \emph{view representation} of $View_v$ as:
\begin{equation}\label{equation:viewRepr}
\mathbf{w}_v = [\mathbf{m}_v^{\mathrm{T}}, \boldsymbol{\sigma}_v^{\mathrm{T}}]^{\mathrm{T}},
\end{equation}
so that the L2 distance of the view representations is now equivalent to the distributional distance, i.e.
$\frac{1}{2}\lVert\mathbf{w}_u - \mathbf{w}_v\rVert^2_2 = d_V(View_u, View_v)^2$.
}

\re{
With the view representation, we can model the generalizable view prototype as a cluster of views.
View clusters can be generalizable and robust,
since slight deviation of viewing condition is allowed in a view cluster,
and hence each view cluster can cover and deal with the potential deviations of a new unseen view.
For example, in a shopping mall,
a newly added camera view facing a passageway may probably find a view cluster consisting of several other passageways views that share similar viewing conditions.
One can use any clustering algorithm according to specific requirements,
and in our method we adopt the simplest K-means clustering, as formulated by:
\begin{equation}\label{equation:viewClustering}
\small
\mathop{\min}_{\{\mathbf{b}_j\}_{j=1}^{J}}f_{vc}=\frac{1}{V} \sum_{j=1}^{J} \sum_{v \in {\mathcal{B}_j}} \lVert\mathbf{w}_v - \mathbf{b}_j\rVert^2_2,
\end{equation}
where
$V$ is the number of views, $J$ is the number of view clusters,
$\mathbf{b}_j$ denotes the centroid of the $j$-th cluster and
$\mathcal{B}_j = \{ v | \mathbf{w}_v \in j$-th cluster$\}$.
}


\re{
After obtaining $J$ view clusters, we regard the view clusters as the \emph{views} for training DECAMEL.
After training, if a new view comes,
we first assign this view to the most similar view cluster and use its feature transformation.
We refer to this method as DECAMEL with \textbf{V}iew \textbf{C}lustering (\emph{DECAMEL$_{VC}$}).
We summarize DECAMEL$_{VC}$ in Algorithm \ref{AlgDECAMELVC}.
By DECAMEL$_{VC}$, the learned feature transformations can naturally generalize to unseen views.
}

\re{
\vspace{0.1cm}
\noindent
\textbf{Scalability}. Solving Eq. (\ref{equation:eigen1}) requires eigen decomposition
whose computational complexity is $O((Vd)^3)$ where $V$ is the number of views
and $d$ is the feature dimension which is constant.
In the view-extendable setting,
we can pre-define the number of generalizable view prototypes $J$ ($J\ll V$),
leading to a constant computational complexity of the eigen decomposition in Eq. (\ref{equation:eigen1}).
In the conventional Re-ID setting where training views and testing views come from the same pool,
we can perform view clustering (and pre-define $J$)
\emph{only} for CAMEL (i.e. metric initialization).
Then, the $i$-th ($1\leq i \leq J$) learned transformation is used to initialize all the view-specific transformations
that belong to the views in the $i$-th view cluster (so that in the joint learning of DECAMEL we still learn $V$ transformations).
We refer to this method as DECAMEL \emph{Initialized} with View Clustering (\emph{DECAMEL$_{IVC}$}).
We find that using a relatively small $J$ for DECAMEL$_{IVC}$ could achieve very close performance
as DECAMEL. The experimental result is shown in the supplementary material.
}

\re{
Finally we also note that DECAMEL$_{VC}$ is a generalization of both DECAMEL and its symmetric version
in the conventional Re-ID setting.
If we set $J = V$, the method is equivalent to original DECAMEL.
On the other hand, if we set $J = 1$, the method degrade to learning a universal feature transformation.
In this sense, DECAMEL$_{VC}$ allows us to flexibly control the generalizability vs. ability to accurately model each specific viewing conditions,
according to the given application scenario.
}

\section{Insight Understanding}\label{section:insight}
In this section, we show visual results to provide intuitive perceptions on
the mechanism of our framework.
We first illustrate that the learned asymmetric metric can alleviate the view-specific bias in the original/initialized feature space
more easily than a symmetric one.
Then, we illustrate how DECAMEL progressively learns a better cross-view cluster structure based on the learned asymmetric metric
by joint learning, and how it mines the potential cross-view discriminative information.

\subsection{Asymmetric Metric Alleviates Views-specific Bias}\label{section:asymmetricAligns}
In Figure \ref{figure:distribution}, we show three cross-view data distributions,
which are in the original/initialized feature space, the shared space learned by a symmetric version of CAMEL
and the shared space learned by CAMEL, respectively.

We can see from Figure \ref{figure:originalDistribution} that in the original feature space, there is severe view-specific bias.
One can easily draw a borderline to separate the data from different views.
Then, in Figure \ref{figure:symmetricDistribution},
the view-specific bias is still severe in the shared space found by the learned symmetric metric (view-generic transformation),
mainly because only some view-generic rotations and translations are taken for both views.
In contrast, in Figure \ref{figure:asymmetricDistribution}, in the shared space learned by CAMEL,
the view-specific bias is alleviated so that data points from two views are much more overlapped with each other.
One of the main reasons is that the learned view-specific transformations provide more flexibility
to facilitate alleviating the bias.

\subsection{DECAMEL Learns Cross-view Cluster Structure}\label{section:DECAMELEmbeds}
In Figure \ref{figure:FM}, we further show the cross-view distributions in both the feature space and the shared space learned by DECAMEL in different stages.
By examining the distributions through stages, we can obtain an intuitive understanding for DECAMEL,
in terms of how it progressively learns a better cross-view cluster structure.


\vspace{0.1cm}
\noindent \textbf{Initialization}.
The metric initialization (i.e., CAMEL) learns a preliminary cross-view cluster structure.
We first look at the leftmost column in Figure \ref{figure:FM}.
We can see that in the original feature space (the leftmost figure in the upper row),
the view-specific bias is severe as we have seen in the last subsection (data points are a subset of Figure \ref{figure:originalDistribution}).
In contrast, after the initial metric layer (the leftmost figure in the lower row), the bias is alleviated (data points are a subset of Figure \ref{figure:asymmetricDistribution}).
Then, by comparing them, we can see that
CAMEL learns a preliminary cross-view cluster structure, i.e., the cross-view data points representing the same identity (color)
roughly get closer to each other.


\vspace{0.1cm}
\noindent \textbf{Joint learning}.
The joint learning facilitates learning a better cross-view cluster structure.
The upper row in Figure \ref{figure:FM} shows the feature distributions in several stages.
Through them, we can find that the feature extractor network is guided by the learned cross-view cluster structure
to improve the feature representation, and in the convergence stage the view-specific bias is alleviated.
This shows that the asymmetric metric is embedded into the whole network, as we discussed in Remark 2.
This is mutually helpful in learning the asymmetric metric.
The lower row in Figure \ref{figure:FM} shows the corresponding distributions after asymmetric metric layer.
Through them, we can find that the asymmetric metric gradually learns a better cross-view cluster structure.
Take the purple and blue identities for example.
In the initial shared space (the leftmost figure in the lower row) there is around $1/3$ data points of them overlapped with each other.
However, along with the joint learning procedure, they are getting more and more compact, and finally there is nearly no overlap between them in the convergence stage.

\vspace{0.1cm}
\noindent \textbf{Mining potential discriminative information}.
Finally, by comparing the initial feature distribution (the leftmost figure in the upper row) with the convergence distribution (the rightmost figure in the lower row),
we can find that through the joint learning of DECAMEL,
a better cross-view cluster structure is learned.
Since the view-specific bias has been alleviated by the asymmetric metric, the cross-view data points belonging to the same identity can get closer to each other in the convergence stage,
rather than entangling and overlapping with other identities in the initialized feature space.
Therefore, during this learning procedure DECAMEL is attempting to mine the potential cross-view discriminative information.
We will further show a quantitative result on evaluating the learned cross-view cluster structures in Sec. \ref{section:structure},
which experimentally validates that DECAMEL learns a better cross-view cluster structure compared to using a symmetric metric.

\section{Experiments}\label{section:experiments}
In this section we compare the performance of DECAMEL with other related unsupervised models to show the effectiveness.
Then, we perform experimental validations and analysis to provide further comprehensive understanding of our framework.
\subsection{Datasets}\label{section:datasets}

\begin{figure}[!t]
\centering
\subfigure[]{
\includegraphics[width=0.11\linewidth]{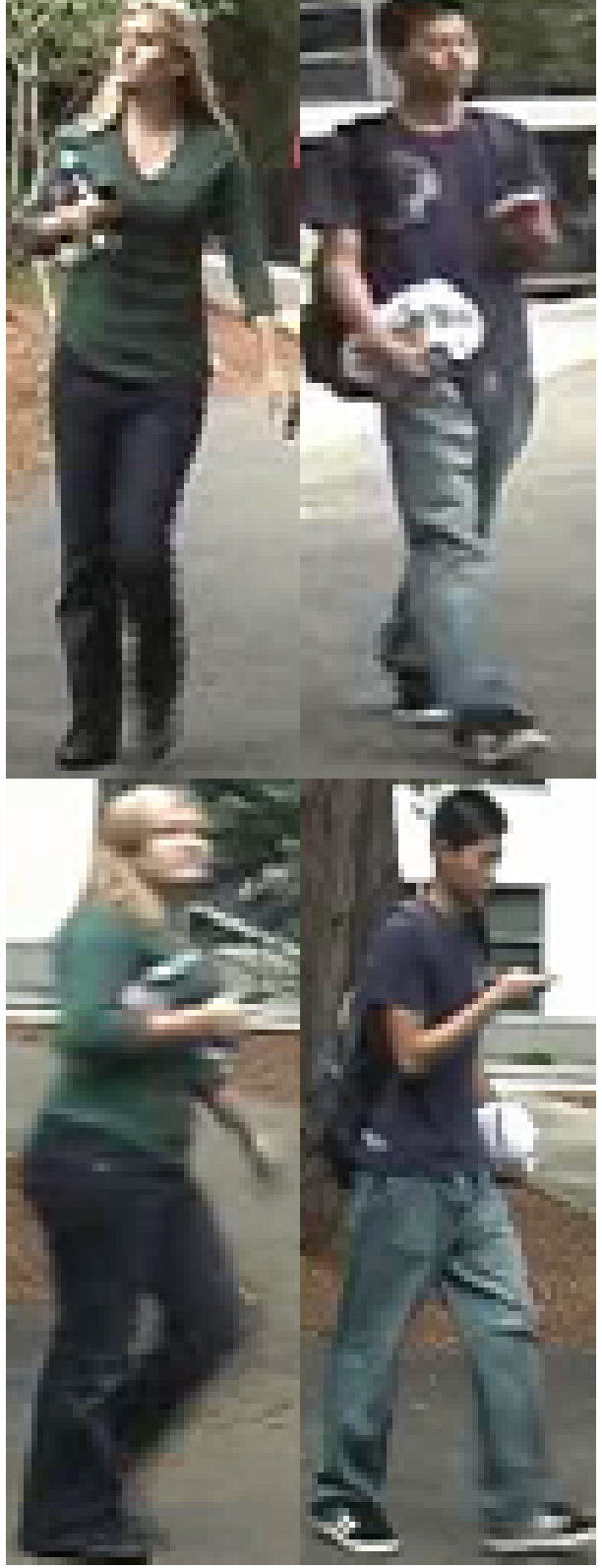}
}
\subfigure[\label{figure:CUHK01}]{
\includegraphics[width=0.11\linewidth]{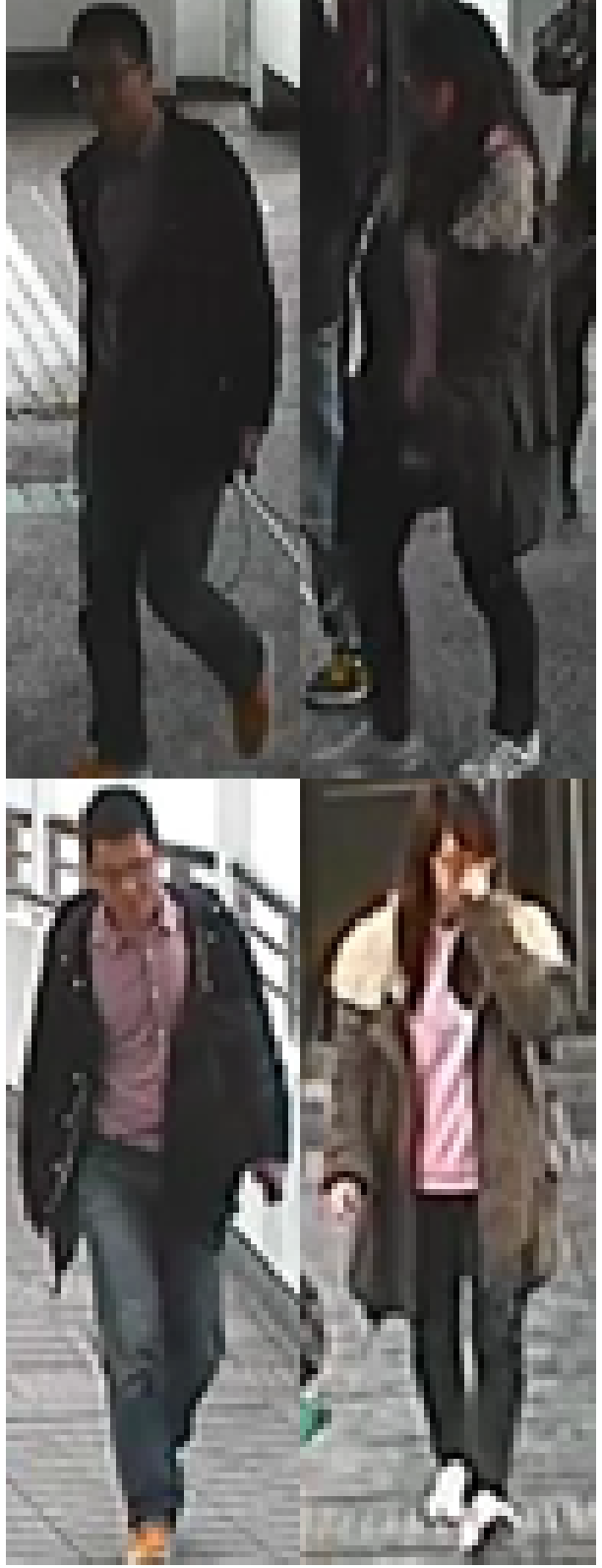}
}
\subfigure[]{
\includegraphics[width=0.11\linewidth]{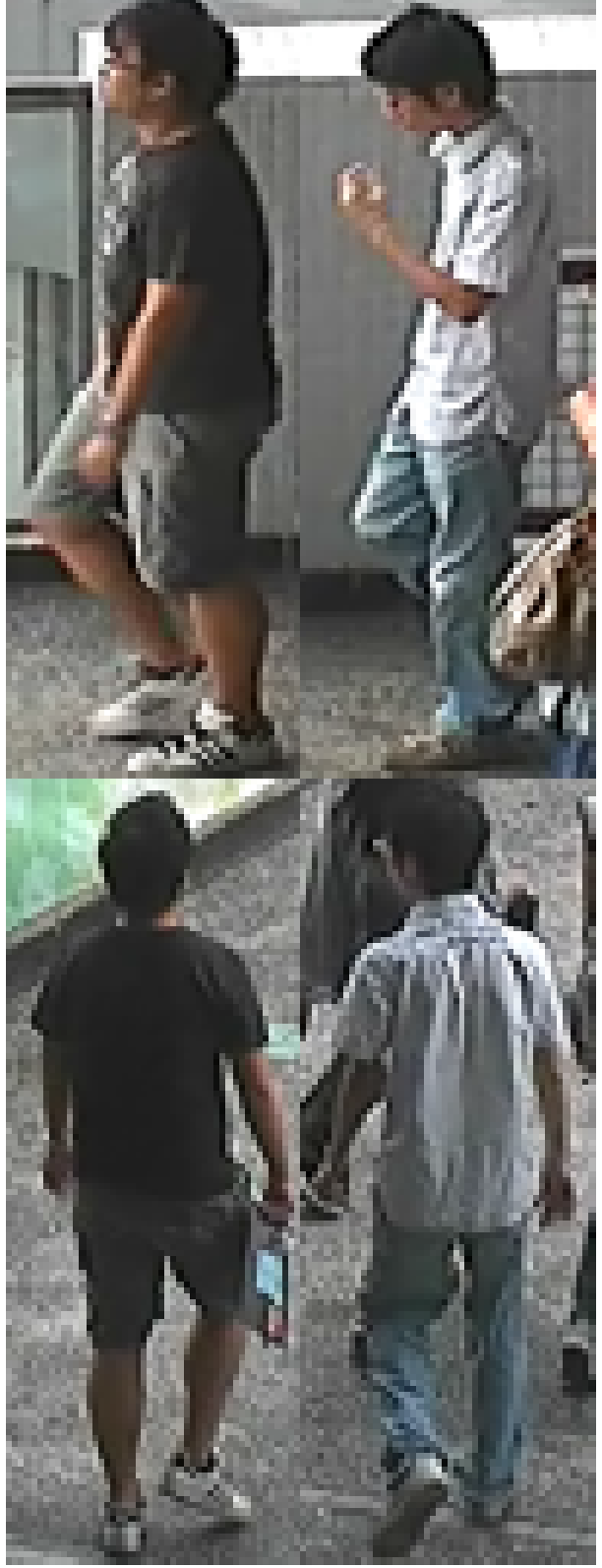}
}
\subfigure[\label{figure:SYSU}]{
\includegraphics[width=0.11\linewidth]{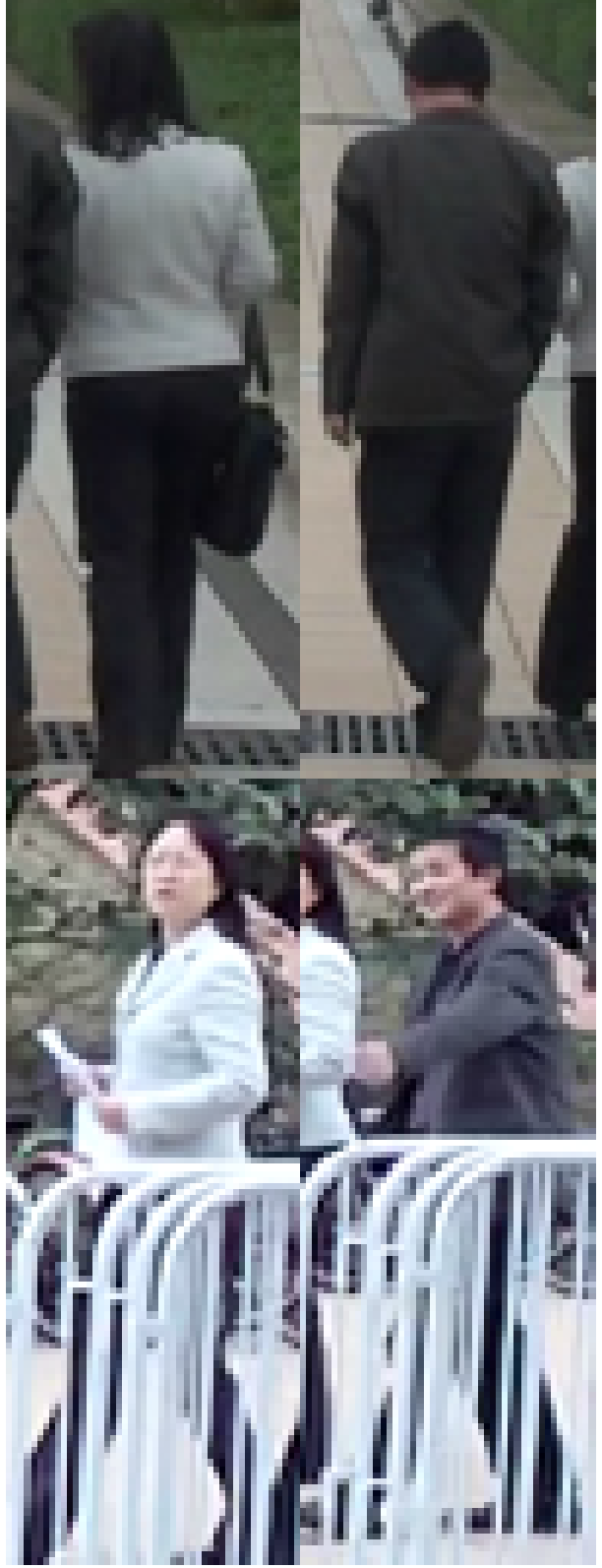}
}
\subfigure[]{
\includegraphics[width=0.11\linewidth]{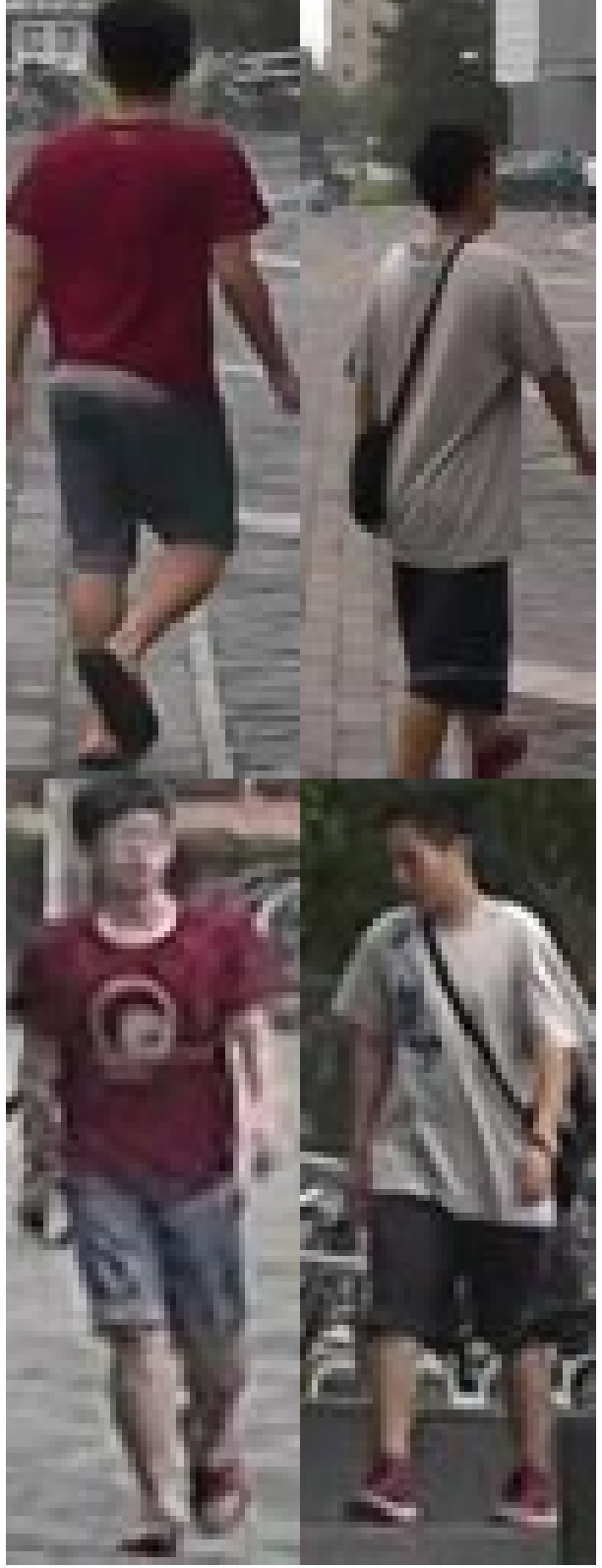}
}
\subfigure[]{
\includegraphics[width=0.11\linewidth]{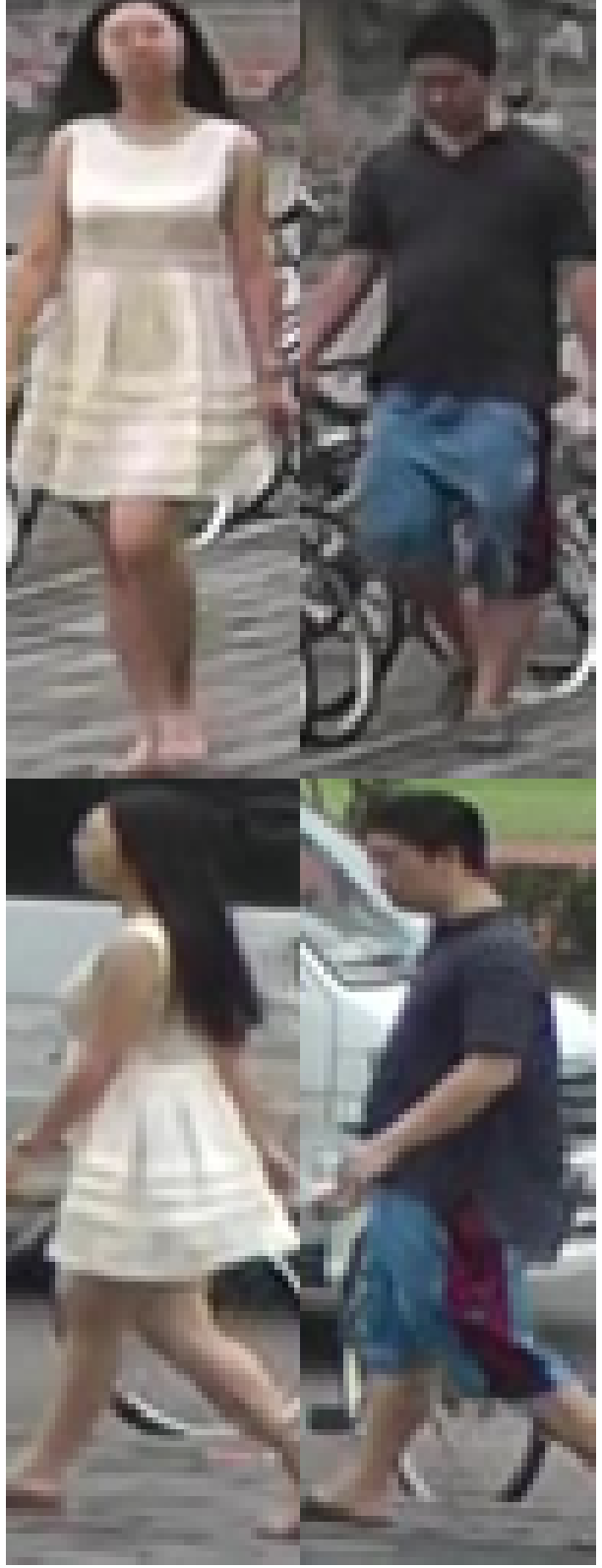}
}
\subfigure[]{
\includegraphics[width=0.11\linewidth]{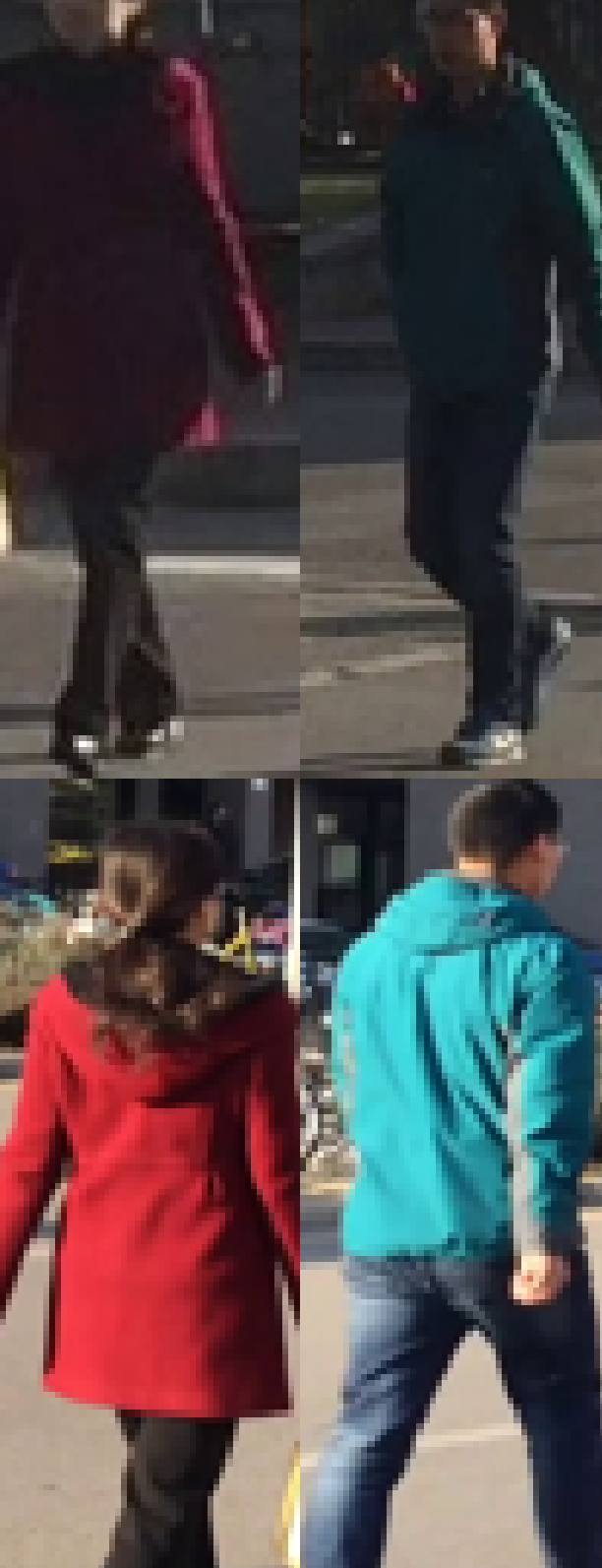}
}
\vspace{-0.4cm}
\caption{\label{FigDatasets}Samples of the datasets. Every two images in
a column are from one identity across two disjoint camera views.
(a) VIPeR (b) CUHK01 (c) CUHK03 (d) SYSU (e) Market (f) ExMarket (g) MSMT17.}
\end{figure}

\begin{table}[!t]
\renewcommand{\arraystretch}{1.1}
\caption{Overview of dataset scales. ``\#'' means ``the number of''.}
\vspace{-0.4cm}
\label{TableDatasets}
\scriptsize
\centering
\begin{tabular}{
>{\centering\arraybackslash}p{1.05cm}|
>{\centering\arraybackslash}p{0.45cm}
>{\centering\arraybackslash}p{0.65cm}
>{\centering\arraybackslash}p{0.65cm}
>{\centering\arraybackslash}p{0.6cm}
>{\centering\arraybackslash}p{0.65cm}
>{\centering\arraybackslash}p{0.65cm}
>{\centering\arraybackslash}p{0.7cm}}
\hline
Dataset      & VIPeR & CUHK01 & CUHK03 & SYSU  & Market & ExMarket & MSMT17 \\
\hline
\# Samples & 1,264 & 3,884 & 13,164 & 24,448 & 32,668 & 236,696 & 126,441 \\
\# Views & 2     & 2     & 6     & 2     & 6     & 6  & 15\\
\hline
\end{tabular}%
\end{table}

We conduct our comparisons
on \re{seven} datasets, whose scales vary from hundreds to hundreds of thousands.
In particular,
since unsupervised models are more meaningful when the scale of problem
is larger, our experiments are conducted on relatively bigger datasets
except VIPeR \cite{VIPER} which is small but widely used.
Various degrees of viewing condition variation can be observed in all these datasets (see Figure \ref{FigDatasets}).
A brief overview of the dataset scales can be found in Table \ref{TableDatasets}.

\vspace{0.05cm}

\noindent \textbf{The VIPeR dataset} \cite{VIPER} contains $1264$ images,
where every two images are captured for each identity from two camera views.

\vspace{0.05cm}

\noindent \textbf{The CUHK01 dataset} \cite{CUHK01} contains $3,884$ images of
971 identities captured from
two disjoint views.

\vspace{0.05cm}

\noindent \textbf{The CUHK03 dataset} \cite{d1} contains $13,164$ images
of 1,360 pedestrians captured from six surveillance camera views.
Pedestrian images were detected by
a state-of-the-art pedestrian detector.

\vspace{0.05cm}

\noindent \textbf{The SYSU dataset} \cite{2015_TCSVT_ASM} includes $24,448$ RGB images of $502$ persons under two surveillance cameras.
One camera view mainly
captured the frontal or back views of persons, while the other observed mostly
the side views.

\vspace{0.05cm}

\noindent \textbf{The Market-1501 dataset} \cite{2015_ICCV_MARKET} (Market) contains $32,668$ images of $1,501$ pedestrians, each of which was
captured by at most six cameras. All of the images were cropped by a pedestrian
detector. There are some badly-detected samples in this datasets as distractors
as well.

\vspace{0.05cm}

\noindent \textbf{The ExMarket dataset}.
Unsupervised models are more meaningful when
the problem scale is larger due to the difficulty in labelling substantial cross-view data.
In order to evaluate unsupervised Re-ID methods on an even larger scale,
we further combined \textbf{the MARS dataset} \cite{MARS} with
Market. MARS is a video-based Re-ID dataset which contains
$20,715$ tracklets of $1,261$ pedestrians. All the identities from MARS are of a
subset of those from Market.
We then took $20\%$ frames (each one in every five successive frames) from the tracklets
and combined them with Market to obtain an extended version of Market (\textbf{ExMarket}).
The imbalance between the numbers of samples from the $1,261$ persons and other
$240$ persons makes this dataset more challenging and realistic. There are $236,696$ images
in ExMarket in total, and $112,351$ images of them are of training set.

\vspace{0.05cm}

\noindent \textbf{The MSMT17 dataset} \cite{2018_CVPR_PTGAN}
\re{is currently the most large-scale dataset which contains $126,441$ images of $4,101$ persons captured from $15$ camera views
during four days. Extreme lighting variations can be observed across camera views.}

\subsection{Settings}\label{section:settings}

\begin{table*}[!t]
\renewcommand{\arraystretch}{1.1}
\caption{Comparison with related unsupervised models: single-shot (``single'') and multi-shot (``multi'') rank-1 matching rate
and MAP in percentage.
In each column, the best is indicated in {\color{red}\textbf{red}} and the second in {\color{blue}\textbf{blue}}.
}
\vspace{-0.4cm}
\label{table:comparison}
\centering
\begin{tabular}{c|cccccccccc}
\hline
Dataset      & VIPeR & CUHK01 & CUHK01 & CUHK03 & CUHK03 & SYSU & SYSU &  Market    & ExMarket & \re{MSMT17}     \\
\hline
Measure      & single & single & multi & single & multi  & single & multi & multi(MAP) & multi(MAP)& multi(MAP) \\
\hline
DIC \cite{Dic}    & \blue{\textbf{29.94}}  & 49.31 & 52.85 & 27.38  & 36.51  & 21.28 & 28.56 & \blue{\textbf{50.21}}(\blue{\textbf{22.68}}) & \blue{\textbf{52.18}}(\blue{\textbf{21.19}})&\blue{\textbf{22.81}}(\blue{\textbf{7.01}}) \\
ISR \cite{ISR}    & 27.53  & \blue{\textbf{53.17}}      & \blue{\textbf{55.66}}   & \blue{\textbf{31.13}} & \blue{\textbf{38.50}} & \blue{\textbf{23.16}}   & \blue{\textbf{33.77}} & 40.32(14.27) & 42.99(15.74) & 21.50(6.10)   \\
RKSL \cite{RKSL}  & 25.76  & 45.41      & 50.13   & 25.79      & 34.75   &  17.64  & 23.01  & 33.97(11.03) & 34.86(10.40) & 15.41(4.30) \\
SAE \cite{SAE1}   & 20.70  & 45.33      & 49.94   & 21.18      & 30.51      & 18.02   & 24.15   & 42.40(16.23) & 43.97(15.10) & 19.29(5.50) \\
JSTL \cite{d4}             & 25.73  & 46.26  & 50.61  & 24.66     & 33.15  & 19.92 & 25.59    & 44.69(18.36)& 46.41(16.68)& 21.24(6.05) \\
\hline
AML \cite{2007_CVPR_AML}  &23.10& 46.78 & 51.14 & 22.19 & 31.41 & 20.88 & 26.39 & 44.71(18.36) & 46.20(16.22)  & 21.16(6.08)\\
UsNCA \cite{2015_NC_uNCA} &24.27& 47.01 & 51.70 & 19.76 & 29.59 & 21.07 & 27.18 & 45.22(18.91) & 47.03(16.91) & 22.01(6.53) \\
\hline
DECAMEL      &\red{\textbf{34.15}}&\red{\textbf{65.81}}&\red{\textbf{69.00}}& \red{\textbf{38.27}}  & \red{\textbf{45.82}}  &\red{\textbf{36.14}} &\red{\textbf{43.90}}& \red{\textbf{60.24}}(\red{\textbf{32.44}}) & \red{\textbf{62.98}}(\red{\textbf{33.28}}) & \red{\textbf{30.34}}(\red{\textbf{11.13}}) \\
\hline
\end{tabular}%
\end{table*}
%
%
%
%

\begin{figure*}[!t]
\centering
\subfigure[CUHK01]{
\includegraphics[width=0.31\linewidth, height=0.2\linewidth]{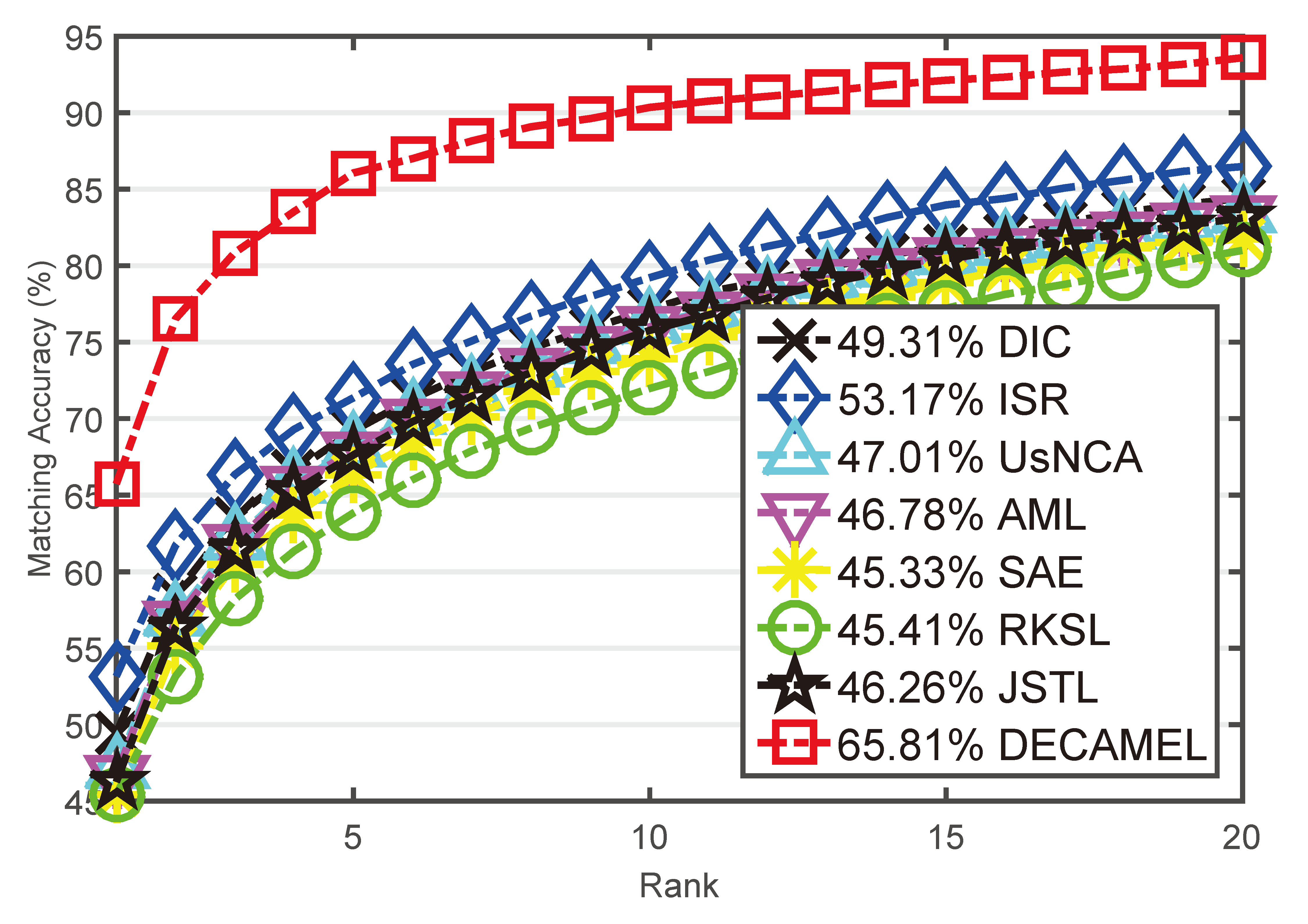}
}
\subfigure[CUHK03]{
\includegraphics[width=0.31\linewidth, height=0.2\linewidth]{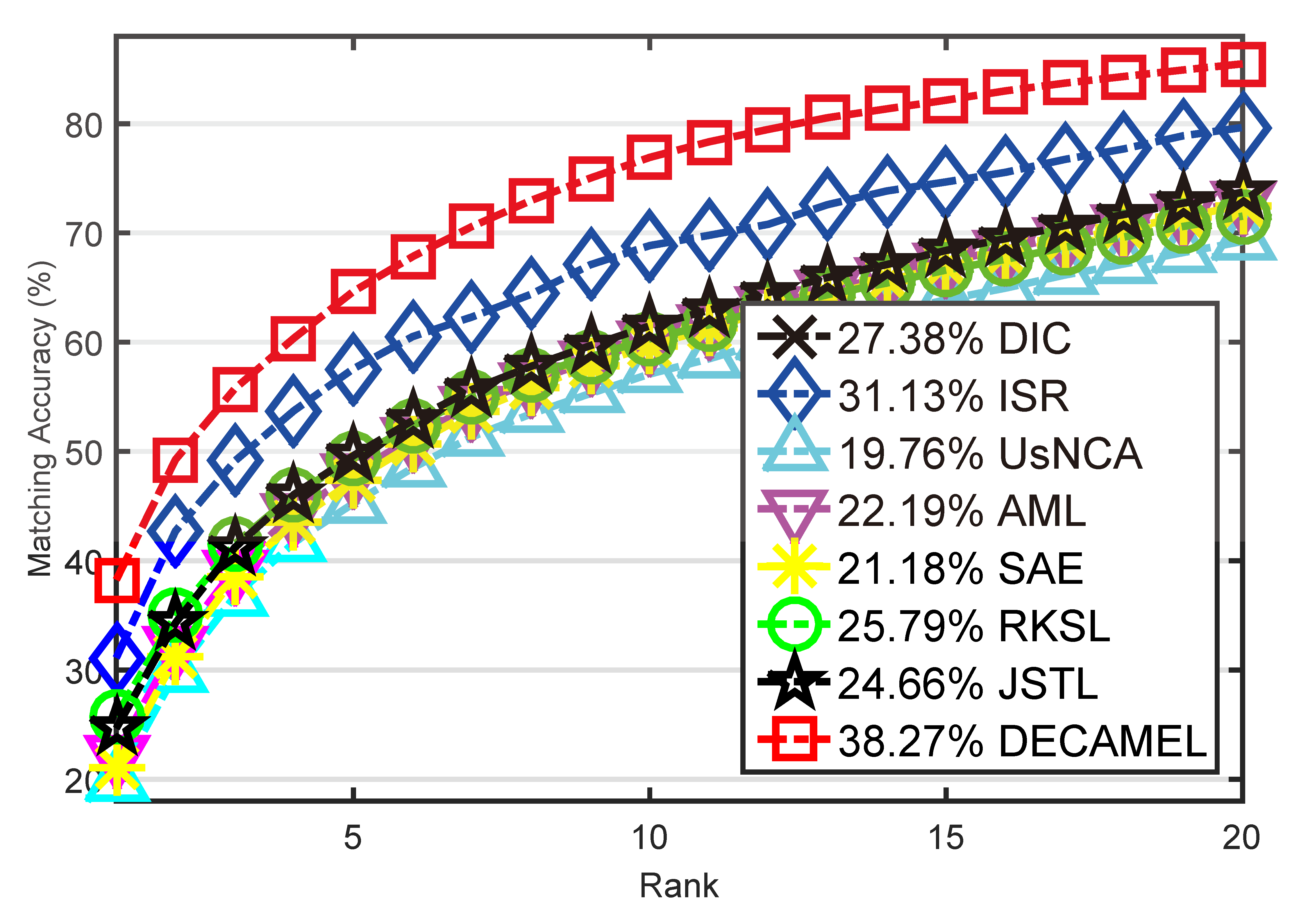}
}
\subfigure[SYSU]{
\includegraphics[width=0.31\linewidth, height=0.2\linewidth]{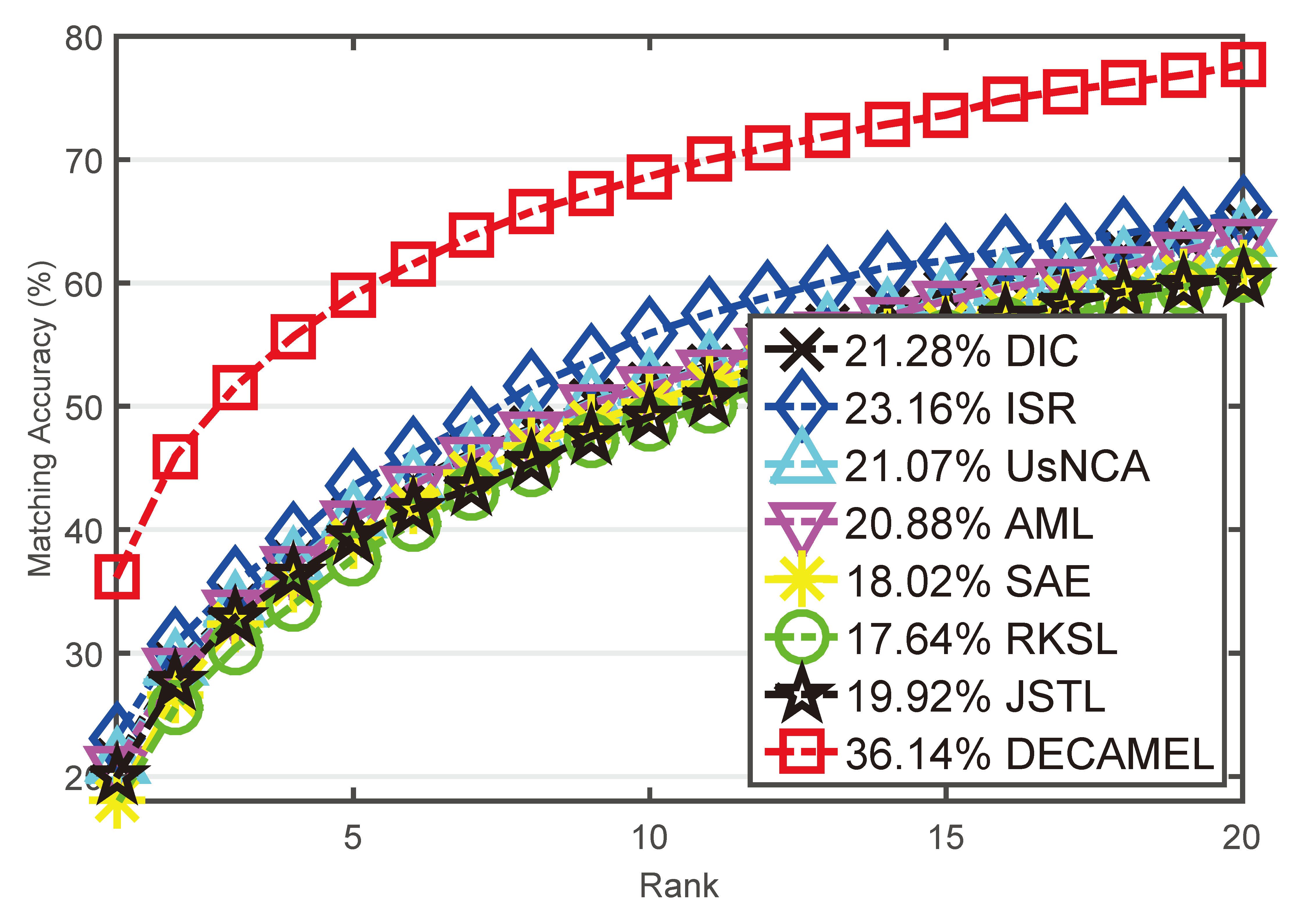}
}
\subfigure[Market]{
\includegraphics[width=0.31\linewidth, height=0.2\linewidth]{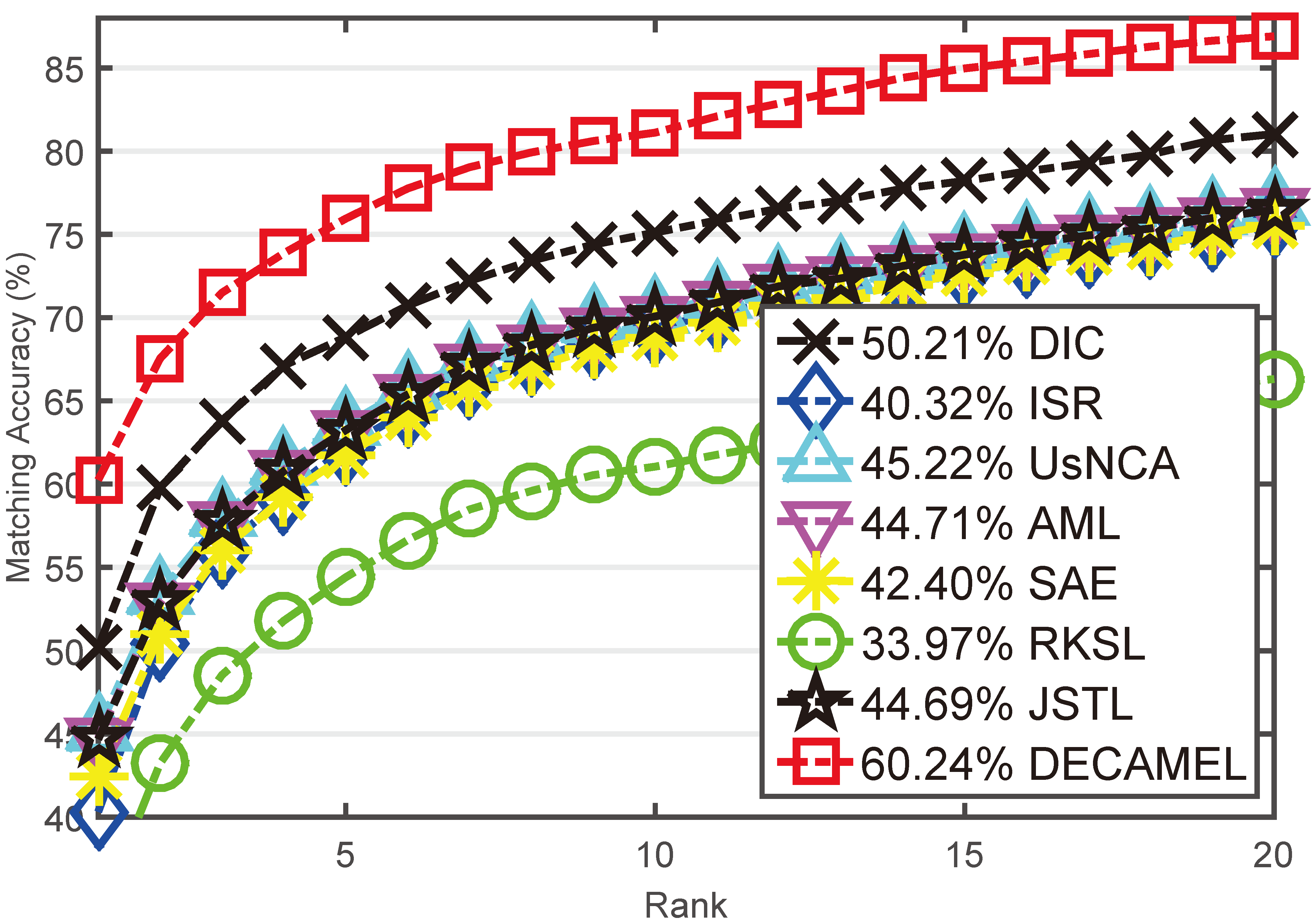}
}
\subfigure[ExMarket]{
\includegraphics[width=0.31\linewidth, height=0.2\linewidth]{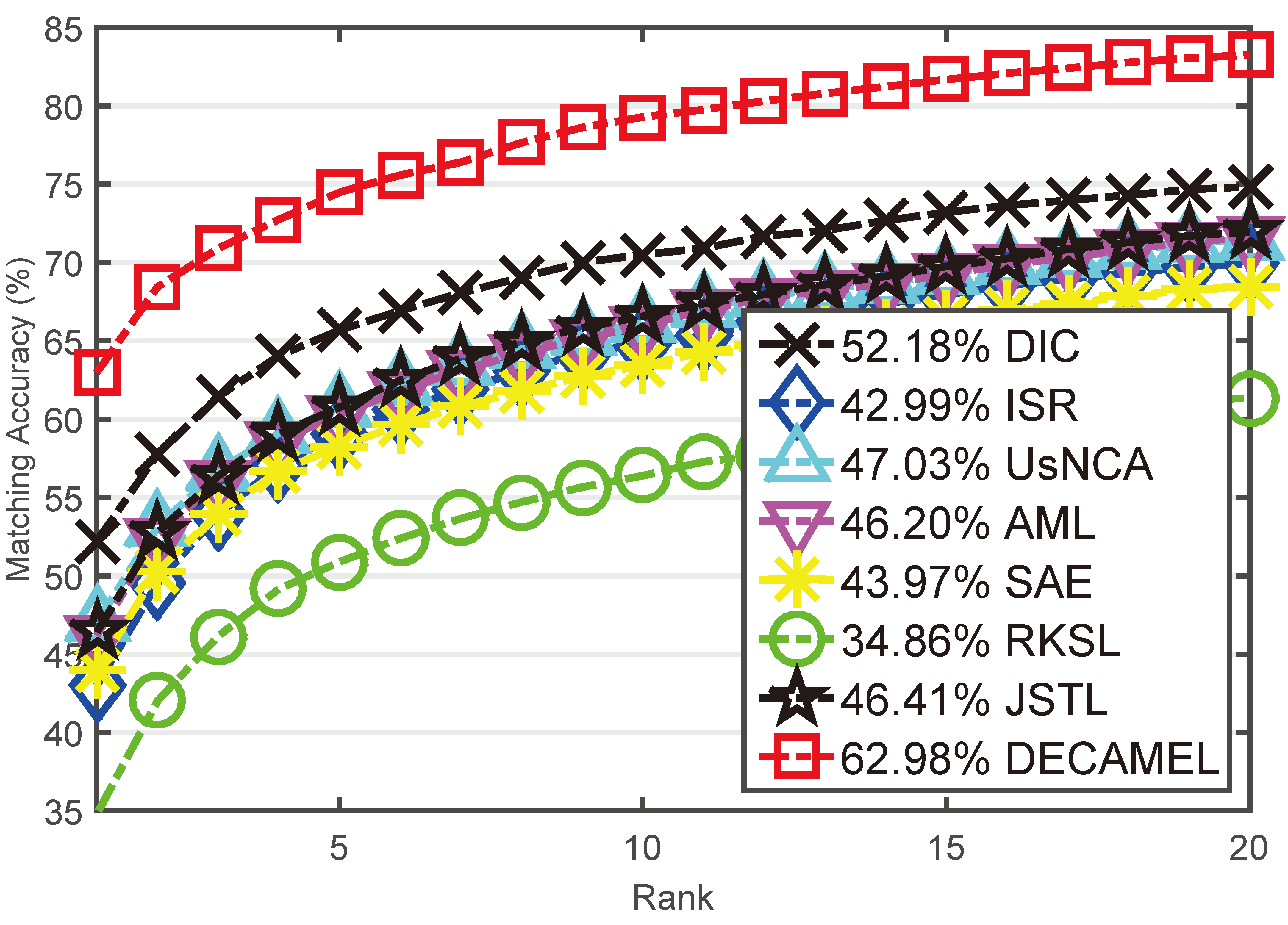}
}
\subfigure[MSMT17]{
\includegraphics[width=0.31\linewidth, height=0.2\linewidth]{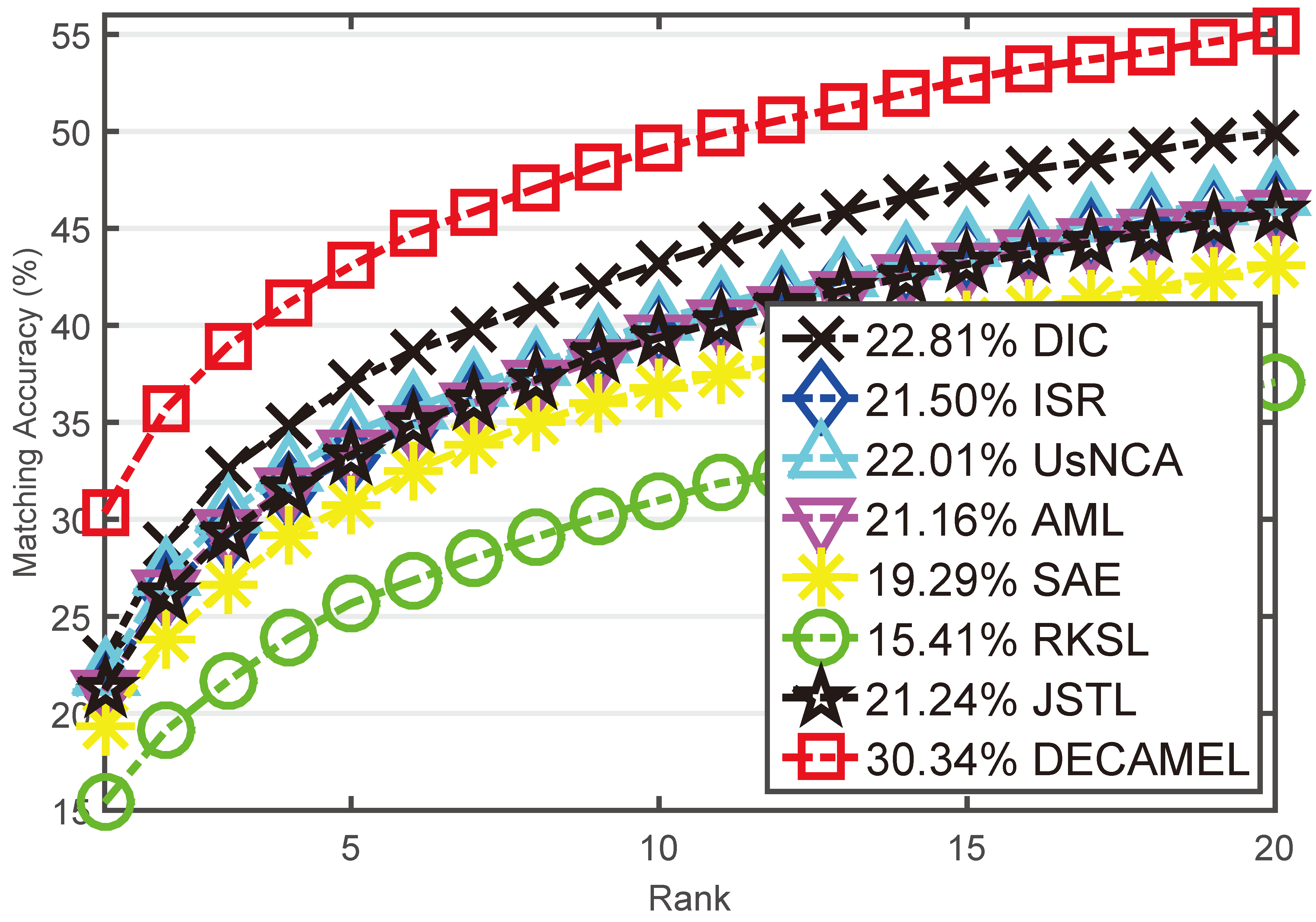}
}
\vspace{-0.4cm}
\caption{CMC curves for comparisons with related unsupervised models.
In each legend, the figure beside the model name is the rank-1 matching rate.
For clarity, we omit VIPeR and show the single-shot results for CUHK01, CUHK03 and SYSU.
}
\label{figure:CMC}
\end{figure*}

\noindent \textbf{Experimental Protocols}.
We follow a widely adopted protocol on VIPeR \cite{m7}, i.e., randomly dividing the images into
two halves, one of which is used as training set and the other as test set. This
procedure is repeated $10$ times to offer an average performance.
This dataset only allows single-shot experiments.

The experimental protocol for CUHK01 was the same as that in \cite{m7}.
We randomly selected images from $485$ persons to form the training set and images from the rest $486$ persons formed the testing set.
The evaluating procedure is repeated $10$ times.
We perform both multi-shot and single-shot experiments.
That is, in the single-shot setting only one image of each gallery person is used for evaluation,
whereas in the multi-shot setting all the images of each gallery person are used.
In both settings, all probe images are used.

The CUHK03 dataset is provided together with its recommended evaluating protocol \cite{d1}.
We follow the provided protocol, where images of $1,260$ persons are chosen as the training set,
and the remainders as testing set.
This procedure is repeated $10$ times.
Both multi-shot and single-shot experiments are conducted.

As for the SYSU dataset, we randomly pick half pedestrians as training set
and the others as testing set.
In the testing stage, we basically follow the protocol as in \cite{2015_TCSVT_ASM}.
That is, we randomly choose one and three images of each pedestrian as gallery for single-shot and multi-shot experiments, respectively.
We repeat the testing procedure by $10$ times.

Market is somewhat different from others. The evaluation protocol is also
provided along with the dataset \cite{2015_ICCV_MARKET}. Since images of one person
come from at most six views, the provided protocol does not adopt the single-shot setting.
Instead, the protocol adopts the multi-shot setting and requires both the cumulative matching characteristic (CMC) and
the mean average precision (MAP) \cite{2015_ICCV_MARKET}.
The protocol of ExMarket is identical to Market since the identities from both datasets are
completely the same as we mentioned above.
\re{For MSMT17 \cite{2018_CVPR_PTGAN} we also use the provided protocol.}

\noindent \textbf{Implementation Details}.
We adopt the 56-layer ResNet \cite{2016_CVPR_ResNet} as the feature extractor network
where the dimension is 64.
The network is pre-trained using the JSTL pre-training technique proposed in \cite{d4},
\re{which used softmax loss with a concatenation of several Re-ID datasets
including CUHK03 \cite{d1}, CUHK01 \cite{CUHK01}, PRID \cite{PRID},
VIPeR \cite{VIPER}, 3DPeS \cite{3DPES}, i-LIDS \cite{ILIDS} and Shinpuhkan \cite{SHINPUHKAN}.}
We do not exploit fine-tuning or the domain guided dropout proposed in \cite{d4}.
Note that we do not use any label of the target dataset.
For example, when pre-training the feature extractor network for the CUHK03 dataset,
we exclude all the training samples from the CUHK03 dataset,
to guarantee an unsupervised setting \cite{2017_ICCV_asymmetric}.
\re{After pre-training, we remove the last fully-connected layer and take the output of the second-last global average pooling layer \cite{2016_CVPR_ResNet}
as our feature.}
\re{Also note that
for MSMT17 \cite{2018_CVPR_PTGAN}, since it is highly challenging for unsupervised setting due to extreme lighting variations,
we add Market-1501, SYSU and the Duke dataset \cite{Duke1, Duke2} to its pre-training set to improve the baseline feature.}

We set $\lambda = 0.01$, $K = 500$ and fix them for all the datasets in the following comparisons.
We also show a parameter evaluation on these major hyper parameters which are corresponding to certain characteristics of DECAMEL.
We typically set the SGD iterations to $10,000$ and
the learning rate to $0.005$ which is divided by $5$ after $5,000$ iterations.
No weight decay ($L_2$ regularization) \re{is applied}.
The batch size is $216$.
\re{
To guarantee that each batch contains all views,
we first compute the distribution of the numbers of samples in each view in the training set (say we have two views and the distribution is [0.4, 0.6]),
and then we randomly sample 0.4*216=86 images in the first view and 0.6*216=130 images in the second view.
We note that we can also use standard random sampling and this empirically does not affect the performance,
as shown in the supplementary material.
}
The framework is implemented based on the MatConvNet \cite{matconvnet}.

\subsection{Comparison to Related Unsupervised Models}
\noindent
\textbf{Comparison to related unsupervised Re-ID models}.
We first compare DECAMEL with the unsupervised Re-ID models.
For a more fair and comprehensive comparison, we conduct experiments on the seven datasets
for DECAMEL and the code-available related models.
The compared models in the following comparisons adopt the same baseline JSTL feature which is used for initialization in DECAMEL.
We have tuned the hyper parameters for the compared models to adapt to the JSTL feature,
and thus report even better results than the original in literatures \cite{Dic, ISR, RKSL}
(the performances are worse than the original without this tuning procedure).
We use the available code for the sparse dictionary learning model (denoted as DIC) \cite{Dic},
the sparse iterative re-ranking model ISR \cite{ISR},
the CCA-based kernel subspace learning model RKSL \cite{RKSL},
and sparse auto-encoder (SAE) \cite{SAE1}.
We also compare our model with the baseline JSTL feature \cite{d4}, which adopts the Euclidean distance as its metric.
The comparative results are measured by the cumulative characteristic curve (CMC) and the rank-$1$ matching rate of CMC.
We show the matching rate in Table \ref{table:comparison},
and show the CMC in Figure \ref{figure:CMC}.

As reported in Table \ref{table:comparison}, our model outperforms other models on all the
datasets in both settings.
In addition, from Figure \ref{figure:CMC}, our model outperforms
other models by large margins at any rank.
This is partly because DECAMEL explicitly deals with the view-specific bias problem by learning an asymmetric metric.
Note that the improvements are notably significant on CUHK01 and SYSU.
We can see that the cross-view condition variations are particularly severe on these two datasets as shown in Figure \ref{figure:CUHK01} and Figure \ref{figure:SYSU} intuitively.
For example,
the changes of illumination are extremely severe in Figure \ref{figure:CUHK01} and Figure \ref{figure:SYSU},
and the differences between features from the two views
may be caused more by illumination than by identity under such a situation.
In particular, although the CCA-based model RKSL also produces specific feature projections for different views,
it learns the specific feature projections inconsistently, and thus does not deal with the view-specific bias.
Apart from this, DECAMEL (and CAMEL as reported in Table \ref{table:DECAMEL}) learns a compact cross-view cluster structure for mining potential discriminative information while RKSL does not.



\begin{table}[!t]
\caption{Comparison with the state-of-the-art unsupervised Re-ID models reported in literature.
The performance is measured by rank-1 matching rate (\%) in single-shot setting.
``-'' means no reported result.
In each row, the best is indicated in {\color{red}\textbf{red}} and the second in {\color{blue}\textbf{blue}}}
\vspace{-0.4cm}
\label{table:sota}
\centering
\begin{tabular}{c|cccccc}
\hline
Model          & SDALF  &  UDML  & GL  & SDC & GTS   & DECAMEL \\
               &\cite{SDALF}&\cite{UDML}&\cite{GL}& \cite{SDC}&\cite{GTS}& \\
\hline
VIPeR          & 19.9  & 31.5  & \blue{\textbf{33.5}} & 26.7  & 25.2  &\red{\textbf{34.2}} \\
CUHK01& 9.9  & 27.1 &   \blue{\textbf{41.0}}   & 26.6  &  -   & \red{\textbf{65.8}} \\
CUHK03         & 4.9    & - &    \blue{\textbf{30.4}}       & 7.7  &   -  & \red{\textbf{38.3}}  \\
\hline
\end{tabular}%
\vspace{-0.4cm}
\end{table}

%

\vspace{0.1cm}
\noindent \textbf{Comparison to published state-of-the-art results}.
Now we compare our model with the reported results in published literatures,
including the transfer learning model UDML \cite{UDML},
the hand-crafted feature model SDALF \cite{SDALF},
the graph learning model GL \cite{GL} and
the saliency learning model GTS \cite{GTS} and SDC \cite{SDC}.
We show the comparative results in Table \ref{table:sota}.
Note that these models have not been evaluated on SYSU, Market and ExMarket,
so we can only compare with \re{their} reported results on VIPeR, CUHK01 and CUHK03 (all single-shot).
As shown in Table \ref{table:sota}, DECAMEL outperforms these models.

\vspace{0.1cm}
\noindent \textbf{Comparison to clustering-based metric models}.
We also compare with a typical clustering-based metric learning model AML \cite{2007_CVPR_AML},
and a recently proposed one UsNCA \cite{2015_NC_uNCA}.
As reported in Table \ref{table:comparison} and Figure \ref{figure:CMC},
DECAMEL can also achieve notable improvements over them.
A main reason should be that DECAMEL learns an asymmetric metric to address the view-specific bias problem so as to learn a better cross-view cluster structure.
In contrast, the compared clustering-based models do not take into consideration this issue which is particularly important for Re-ID.

\subsection{Further Analysis of DECAMEL}\label{section:furtherModelAnalysis}

In the following, we provide some further experimental validations and analysis to make a more comprehensive understanding
for the framework components and show some significant properties.

\subsubsection{Asymmetric vs. Symmetric Modelling}\label{section:structure}
We first evaluate the asymmetric modelling in our framework.
To this end, we develop a symmetric version of DECAMEL and denote it as DECMEL.
The only difference between them is that DECMEL learns (and embeds) a symmetric metric instead of an asymmetric one.
\modify{We show the comparative results of performances in the upper part of Table \ref{table:CMEL}.
We can see that DECAMEL achieves much higher performances than DECMEL.

To further explore the differences between DECAMEL and DECMEL,
we also develop a natural, reasonable measure to evaluate the learned cross-view cluster structure in terms of mining the cross-view discriminative information.}
The measure is formulated as
\begin{equation}
\small
\begin{aligned}
S &= inter/intra \\
&= (\frac{1}{P(P-1)}\sum_{p \neq q}^{P} \lVert \mathbf{d}_p-\mathbf{d}_q\rVert_2) / (\frac{1}{T}\sum_{p=1}^P \sum_{i\in \mathcal{D}_p}\lVert \mathbf{x}_i - \mathbf{d}_p\rVert_2),
\end{aligned}
\label{equation:structure}
\end{equation}
where $P$ denotes the number of persons, $\mathcal{D}_p$ is a set containing the indexes of all the cross-view images of the $p$-th person,
$\mathbf{d}_p$ denotes the centroid of the $p$-th person, and $T$ denotes the total number of cross-view person images.
In Eq. (\ref{equation:structure}), the numerator measures the inter-person discrepancy and the denominator measures the intra-person discrepancy.
We note that we only use the label information to form the measure to
evaluate how discriminative the learned cross-view cluster structure is for Re-ID,
i.e., the higher the $S$ value is, the more easily we can distinguish different persons.
We show the comparative results in the lower part of Table \ref{table:CMEL}.
We can see in Table \ref{table:CMEL} that DECAMEL has higher $S$ values on all datasets,
\modify{indicating that the asymmetric modelling in our framework helps learn a better cross-view cluster structure to facilitate mining the potential cross-view discriminative information.
This can be one of the main reasons why DECAMEL outperforms DECMEL.}
In the following parameter evaluation we will further explore the behavior of the proposed asymmetric modelling.

\begin{table}[!t]
\renewcommand{\arraystretch}{1.1}
\caption{Evaluation of the asymmetric modelling in our framework.
``DECMEL'' denotes the symmetric version of DECAMEL.
The $S$ value is defined in Eq. (\ref{equation:structure}).
The performance is measured by single-shot (``single'') and multi-shot (``multi'') rank-1 matching rate
and MAP in percentage.
For clarity, we drop VIPeR and the multi-shot results of CUHK01, CUHK03 and SYSU,
which follow a similar pattern to single-shot results.}
\vspace{-0.4cm}
\label{table:CMEL}
\centering
\scriptsize
\begin{tabular}{
>{\centering\arraybackslash}p{1.0cm}
>{\centering\arraybackslash}p{0.7cm}
>{\centering\arraybackslash}p{0.75cm}
>{\centering\arraybackslash}p{0.45cm}
>{\centering\arraybackslash}p{1.05cm}
>{\centering\arraybackslash}p{1.05cm}
>{\centering\arraybackslash}p{1.05cm}}
\hline
Dataset      & CUHK01  & CUHK03 & SYSU &  Market    & ExMarket & MSMT17     \\

\hline
Measure      & single & single  & single & multi(MAP) & multi(MAP) & multi(MAP) \\
\hline
DECMEL        &55.95& 27.86    &25.38&49.94(23.15)& 53.00(26.53) & 23.88(8.01) \\
DECAMEL      &\textbf{65.81}& \textbf{38.27}  &\textbf{36.14}& \textbf{60.24}(\textbf{32.44}) & \textbf{62.98}(\textbf{33.28})  & \textbf{30.34}(\textbf{11.13})\\
\hline
Measure      & $S$ & $S$ &$S$ &$S$ & $S$ & $S$ \\
\hline
DECMEL        &2.13& 1.44    &1.61&1.79&1.82  & 1.33 \\
DECAMEL      &\textbf{2.36}& \textbf{1.87} &\textbf{1.88}&\textbf{1.93}& \textbf{1.97} & \textbf{1.42} \\

\hline
\end{tabular}%
\vspace{-0.2cm}
\end{table}

\subsubsection{Parameter Evaluation}\label{section:parameterDiscussion}

\begin{figure*}[!t]
\centering
\subfigure[CUHK03\label{figure:lambdaCUHK03}]{
\includegraphics[width=0.22\linewidth]{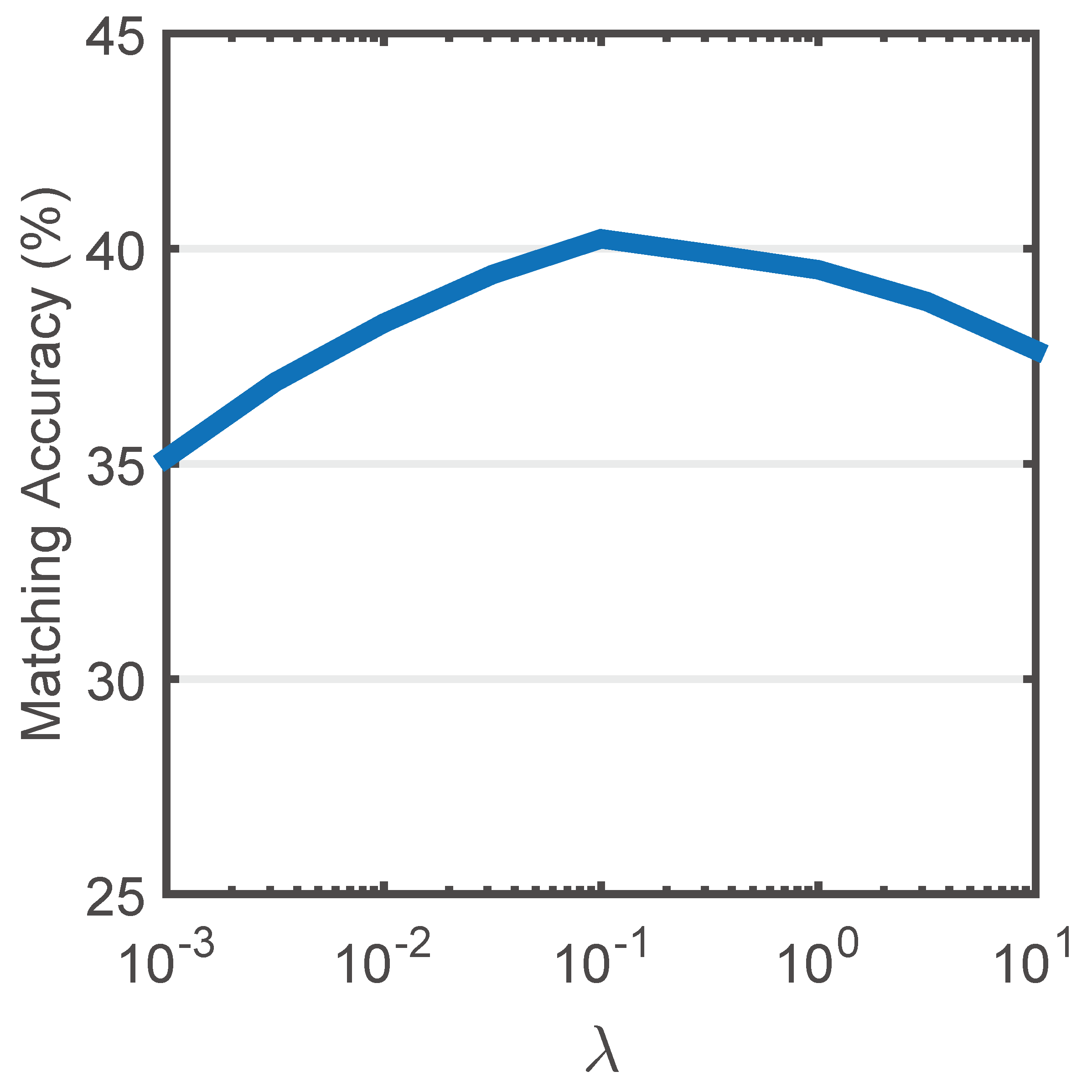}
}
\subfigure[SYSU\label{figure:lambdaSYSU}]{
\includegraphics[width=0.22\linewidth]{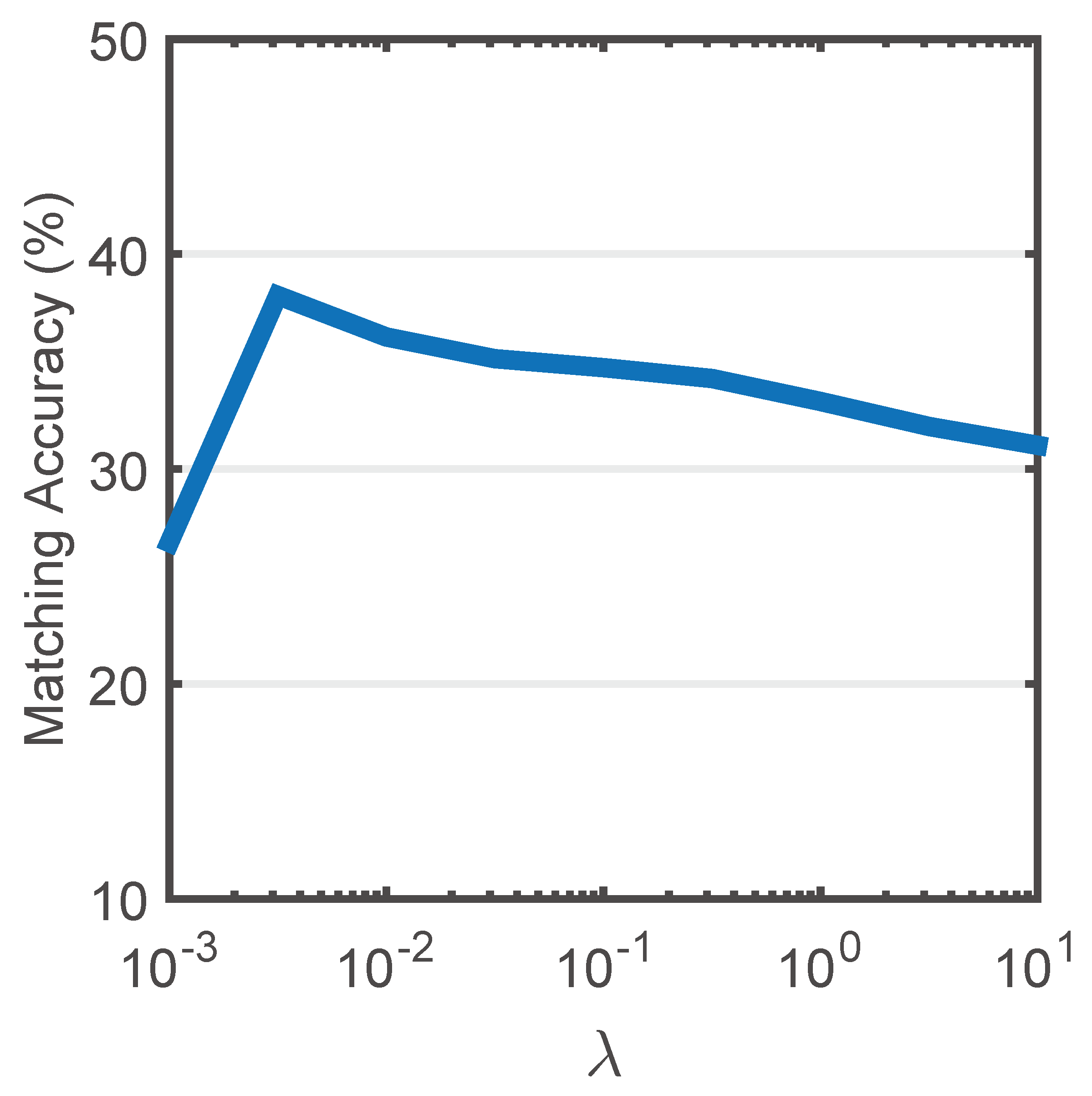}
}
\subfigure[ExMarket\label{figure:lambdaExMarket}]{
\includegraphics[width=0.22\linewidth]{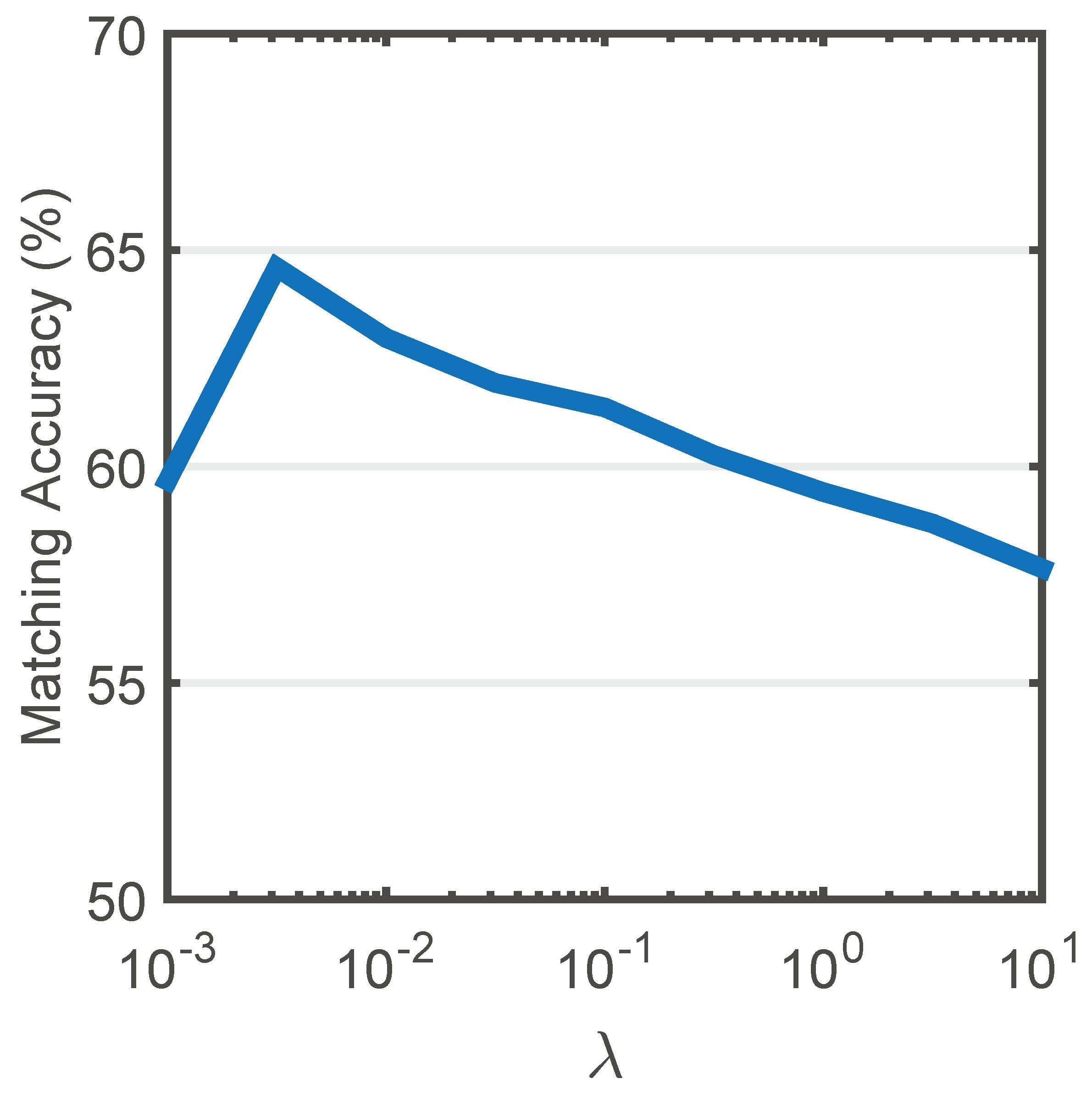}
}
\subfigure[Collapsed distribution\label{figure:collapseDistribution}]{
\includegraphics[width=0.22\linewidth, height=0.22\linewidth]{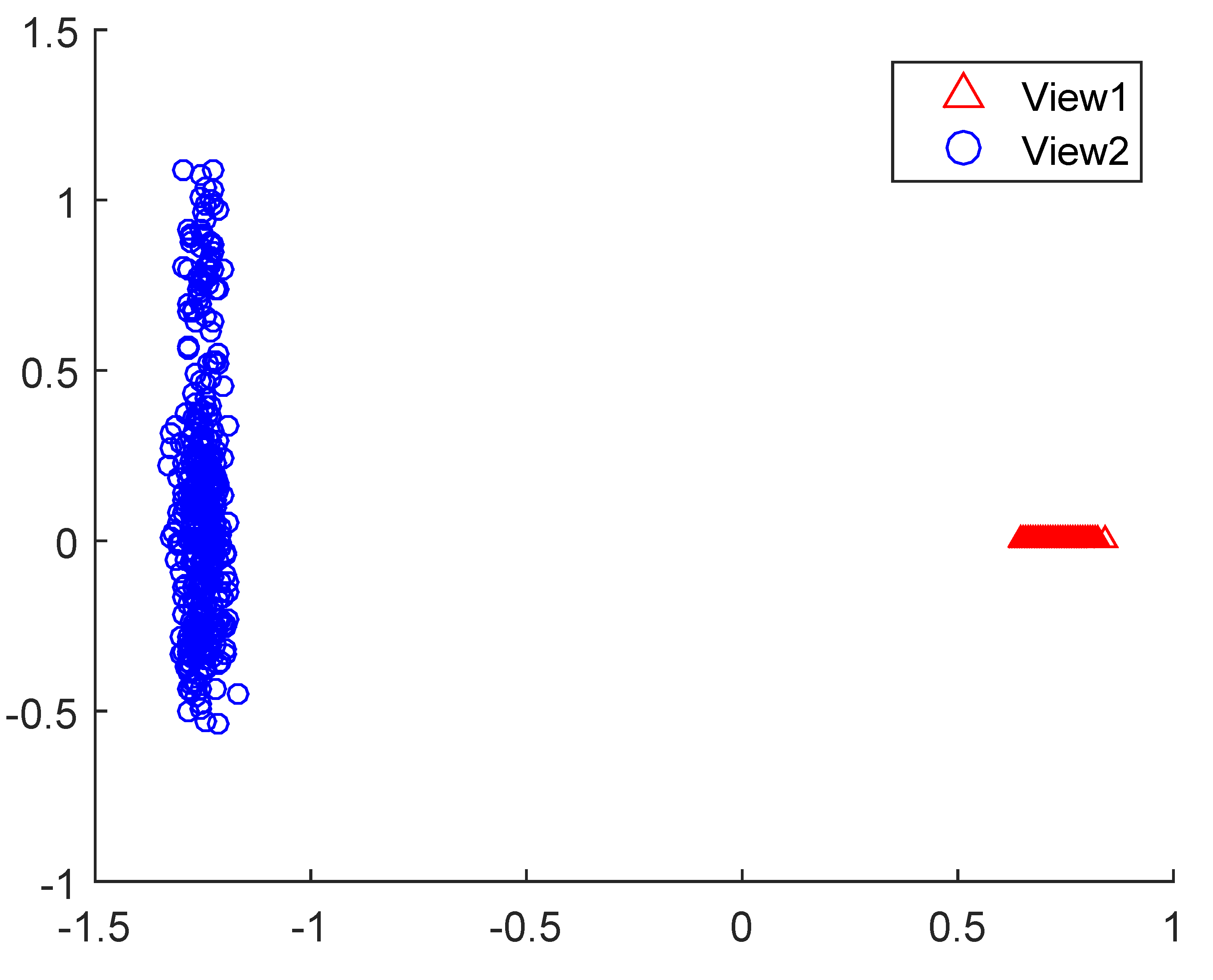}
}
\vspace{-0.4cm}
\caption{(a)-(c) Matching rate vs. $\lambda$ on the three large-scale datasets.
Similar observations can be made on other datasets.
(d) The cross-view distribution without cross-view consistency regularization.
Data is from the SYSU dataset, and we performed PCA for visualization as in Figure \ref{figure:distribution}.
}
\label{figure:Klambda}
\end{figure*}

In the comparisons we fix the major hyper parameters,
and here we discuss the behaviors of the major hyper parameters
to have a better understanding of our framework.
The cores of our framework are asymmetric modelling and cross-view clustering.
They are characterized mainly by $\lambda$ (the cross-view consistency regularizer) and $K$ (the number of clusters), respectively.

\vspace{0.1cm}

\noindent \textbf{Evaluation of $\boldsymbol{\lambda}$: Characteristic of Asymmetric Modelling}.
The cross-view consistency regularizer, $\lambda$, controls the degree of asymmetric modelling.
When $\lambda$ is larger, the larger penalty enforces the discrepancy between any pair of projection basis to be smaller,
and thus the asymmetric modelling will become more symmetric.
We show the matching rate as a function of $\lambda$ in Figures \ref{figure:lambdaCUHK03}, \ref{figure:lambdaSYSU} and \ref{figure:lambdaExMarket}.
When $\lambda$ is in a median range,
the performance is relatively stable.
When $\lambda$ is too large,
the matching rate will drop.
In fact, in the extreme case when $\lambda$ goes to infinity, it is equivalent to the symmetric version.
This shows that the asymmetric modelling is very significant in our framework.

On the other hand, when $\lambda$ is too small, the matching rate also drops.
To understand the reason behind, we examine the extreme case when $\lambda = 0$.
This is equivalent to taking out the cross-view consistency regularization in Eq. (\ref{equation:f_obj1}) and Eq. (\ref{equation:loss}).
In this extreme case, the model actually fails to learn.
To reveal the underlying reason,
we show the cross-view data distribution in the learned shared space without the regularization in Figure \ref{figure:collapseDistribution}.
We find that the distribution collapses roughly to two lines.
This is because the intrinsic consistency across the distributions of different views
is not preserved by the arbitrarily different transformations.
Thus, without the cross-view consistency regularization,
the learned transformations become extreme to minimize the objective and produces the collapsed cross-view distribution.
Clearly, such a distribution loses the discriminative information.
This observation shows that the cross-view consistency regularization averts learning a shared space with collapsed cross-view distribution.

\vspace{0.1cm}

\begin{figure}[!t]
\centering
\includegraphics[width=0.6\linewidth, height=0.4\linewidth]{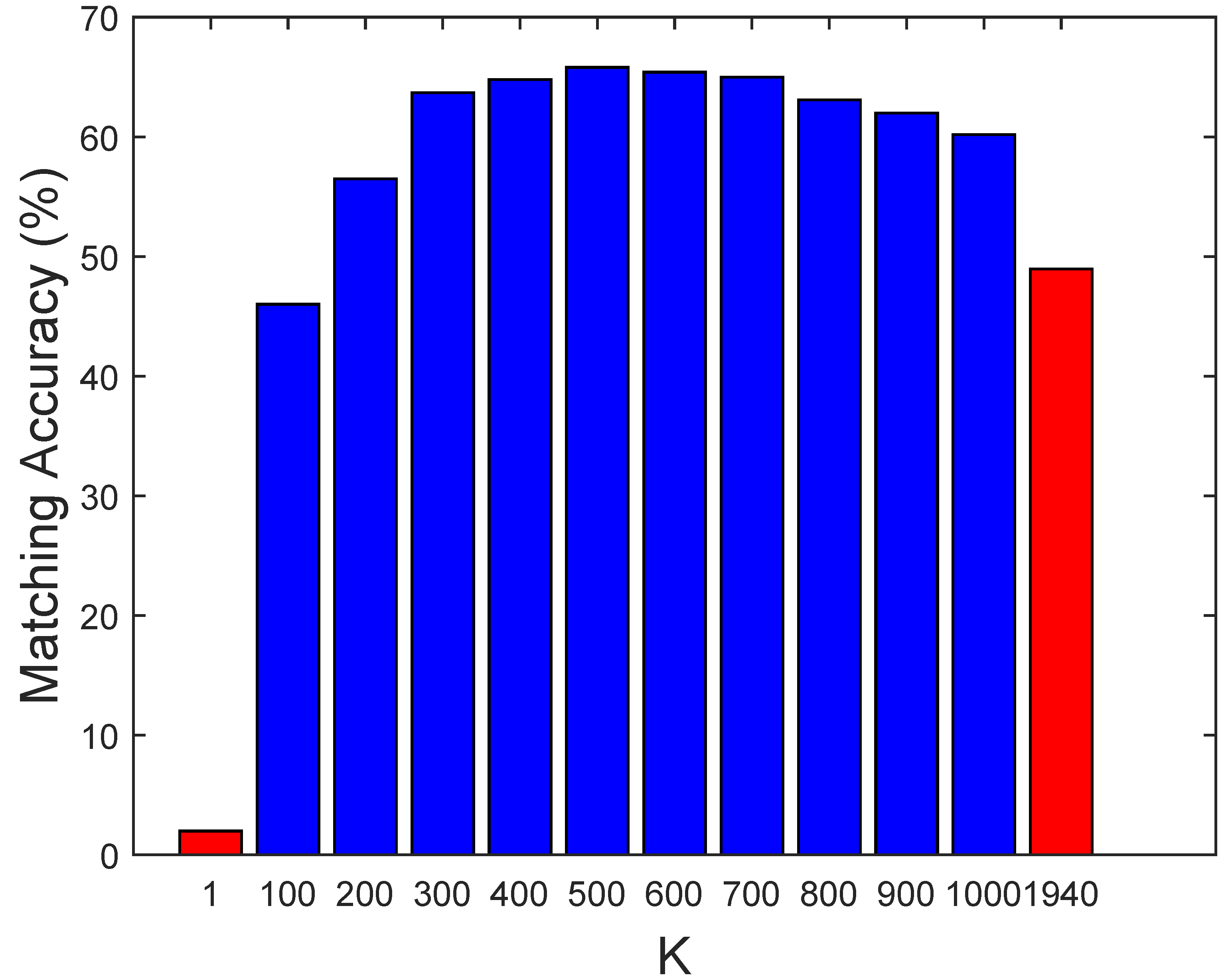}
\vspace{-0.4cm}
\caption{\label{figure:KlambdaCUHK01}Matching rate as a function of $K$ on CUHK01.
$K$ (blue parts) is linearly spaced from $100$ to $1000$.
We show two extreme cases (red parts) when $K = 1$ and $K = 1940$, where $1940$ is the number of total training samples.
Similar observations can be made on other datasets.
}
\vspace{-0.4cm}
\end{figure}

\noindent \textbf{Evaluation of $\boldsymbol{K}$: Characteristic of Asymmetric Metric Clustering}.
$K$ is the number of clusters in DECAMEL.
We show the matching rate as a function of $K$ in Figure \ref{figure:KlambdaCUHK01}.
In the middle blue parts in Figure \ref{figure:KlambdaCUHK01}, when $K$ is set in a median range, e.g., $300$ - $700$, the bars are tightly close to each other.
This shows that to a mild extent,
our framework is robust to $K$.

\begin{figure}[!t]
\centering
\includegraphics[width=0.6\linewidth, height=0.4\linewidth]{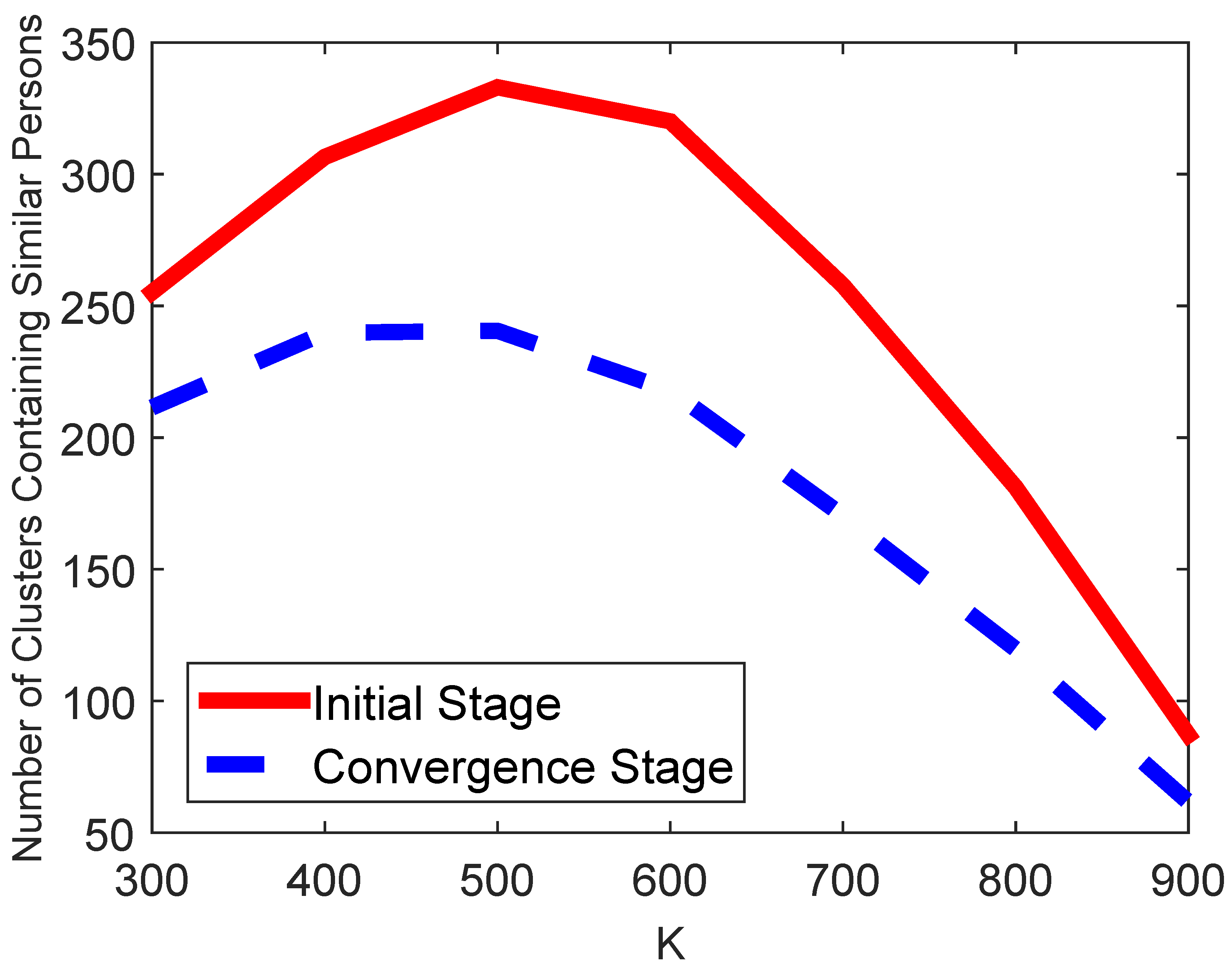}
\vspace{-0.4cm}
\caption{\label{figure:SimilarCUHK01}
Number of clusters containing more than one person at the initial stage (red solid line) and at the convergence stage (blue dashed line)
when $K$ varies on CUHK01.
Similar observations can be made on other datasets.
}
\vspace{-0.4cm}
\end{figure}

To further explore the reason behind, we show in Figure \ref{figure:SimilarCUHK01} the number of clusters which contains more than one person (i.e., $>1$ persons) when $K$ varies.
From Figure \ref{figure:SimilarCUHK01}, it is found that
\emph{(1)} despite $K$ is varying,
there is always a number of clusters containing more than one person in the initialization stage,
i.e., the cluster results are in fact far from perfect.
And \emph{(2)},
in the convergence stage the numbers are consistently decreased compared to the initialization stage.
This indicates that for $K$ in a median range, the cluster results are improved consistently.
This can be a reason for the mild robustness.

However, on the other hand, we can also find that in Figure \ref{figure:KlambdaCUHK01},
the two extreme cases (red parts) where $K = 1$ and $K = 1940$ lead to performance drop.
In the case when $K = 1$, the model fails to learn, somewhat similar to the situation shown in \ref{figure:collapseDistribution},
because $K = 1$ leads to a collapsed distribution where all the data points are pulled towards a single centroid.

\subsubsection{Quantitative Component-wise Evaluation}\label{section:component_wise}

\begin{table}[!t]
\renewcommand{\arraystretch}{1.1}
\caption{Component-wise evaluation.
Feat$_{init}$ denotes the initialized feature. Feat$_{embd}$ denotes the feature learned by metric embedding.
DECAMEL$_{f}$ denotes the model when asymmetric metric layer is frozen while DECAMEL$_{m}$ denotes when feature extractor is frozen.
Performance is measured by single-shot (``single'') and multi-shot (``multi'') rank-1 matching rate
and MAP in percentage. For clarity, we drop VIPeR and the multi-shot results of CUHK01, CUHK03 and SYSU,
which follow a similar pattern to single-shot results.
}
\vspace{-0.4cm}
\centering
\scriptsize
\label{table:DECAMEL}
\centering
\begin{tabular}{
>{\centering\arraybackslash}p{1.0cm}
>{\centering\arraybackslash}p{0.7cm}
>{\centering\arraybackslash}p{0.75cm}
>{\centering\arraybackslash}p{0.45cm}
>{\centering\arraybackslash}p{1.05cm}
>{\centering\arraybackslash}p{1.05cm}
>{\centering\arraybackslash}p{1.05cm}}
\hline
Dataset       & CUHK01 & CUHK03 & SYSU &  Market & ExMarket  & MSMT17   \\
\hline
Measure       & single & single & single & multi(MAP) & multi(MAP) & multi(MAP) \\
\hline
Feat$_{init}$      & 46.26 & 24.66  & 19.92 & 44.69(18.36) & 46.41(16.68) & 21.24(6.05) \\
Feat$_{embd}$      & \textbf{53.67} & \textbf{33.20} & \textbf{28.29} & \textbf{51.90}(\textbf{25.56}) & \textbf{58.37}(\textbf{28.21}) & \textbf{25.31}(\textbf{8.49}) \\
\hline
CAMEL        &57.30& 31.89   &30.76&54.45(26.31)& 55.88(23.88) & 27.06(9.14) \\
DECAMEL$_{f}$  & 64.17 & 36.71 & 34.74 & 55.97(29.30) & 60.01(31.11) & 27.20(9.78) \\
DECAMEL$_{m}$  & 54.88 & 30.15 & 26.06 & 51.19(23.65) & 54.81(23.09)  & 26.74(8.86) \\
DECAMEL      &\textbf{65.81}& \textbf{38.27}  &\textbf{36.14} &\textbf{60.24}(\textbf{32.44}) & \textbf{62.98}(\textbf{33.28}) & \textbf{30.34}(\textbf{11.13}) \\
\hline
\end{tabular}%
\end{table}


Now we discuss how framework components contribute to DECAMEL.
We first set the learning rate to $0$ for the metric layer to freeze it.
By this way, the framework only learns the feature representation.
We denote this derived model as DECAMEL$_{f}$.
Similarly, we freeze the feature extractor,
so we have DECAMEL$_{m}$.
We also compare DECAMEL with the initialization stage (CAMEL).
Besides, we compare the initialized feature representation (denoted as Feat$_{init}$) with the feature learned by metric embedding, i.e.,
we take the feature alone out of DECAMEL for comparison (denoted as Feat$_{embd}$).
We show the results in Table \ref{table:DECAMEL}.

From Table \ref{table:DECAMEL} we can make three observations:
\emph{(1)} by comparing Feat$_{init}$ with Feat$_{embd}$, we find that Feat$_{embd}$ consistently outperforms Feat$_{init}$.
This shows that the deep asymmetric metric embedding also learns some underlying cross-view discriminative information for the feature representation,
as have been illustrated in Sec. \ref{section:DECAMELEmbeds}.
\emph{(2)} By comparing DECAMEL$_{f}$ and DECAMEL$_{m}$ with DECAMEL,
we find that any single component cannot achieve improvements as DECAMEL.
This shows that the two components of DECAMEL are intrinsically joint and cooperative, and their effects are mutually promoting each other, rather than simply linearly superposed.
\emph{(3)} By comparing CAMEL with DECAMEL, we can find that DECAMEL further provides noticeable improvements over CAMEL.
This suggests the potential of jointly learning feature and metric in unsupervised Re-ID.

\subsubsection{Effect of Metric Initialization}\label{section:metricInitialization}
\re{As Remark 2 reveals, the joint learning of DECAMEL is partially guided by the metric.
We compare our proposed metric initialization (CAMEL) to two standard initialization strategies,
i.e., identity matrix (in the view of distance metric) and randomly initialized matrix using the Xavier initialization \cite{2014_ACMMM_caffe}
(in the view of fully-connected layer),
and we denote their results as DECAMEL$_i$ and DECAMEL$_r$, respectively.
As shown in Table \ref{table:init}, DECAMEL outperforms both of them.
This is because in our learning algorithm,
the metric initialization method CAMEL can learn an asymmetric distance metric which captures the cross-view person appearance variations,
providing a cross-view discriminative initialization suitable for the unsupervised feature learning, and therefore achieve superior performance.}

\begin{table}[!t]
\caption{Evaluation of different initialization strategies: single-shot (``single'') and multi-shot (``multi'') rank-1 matching rate
and MAP in percentage. ``DECAMEL$_i$'' and ``DECAMEL$_r$'' denote
DECAMEL initialized by identity matrix and random matrix \cite{2014_ACMMM_caffe}, respectively, rather than CAMEL.
For clarity, we drop VIPeR and the multi-shot results of CUHK01, CUHK03 and SYSU,
which follow a similar pattern to single-shot results.}
\label{table:init}
\vspace{-0.4cm}
\centering
\scriptsize
\begin{tabular}{
>{\centering\arraybackslash}p{1.0cm}
>{\centering\arraybackslash}p{0.7cm}
>{\centering\arraybackslash}p{0.75cm}
>{\centering\arraybackslash}p{0.45cm}
>{\centering\arraybackslash}p{1.05cm}
>{\centering\arraybackslash}p{1.05cm}
>{\centering\arraybackslash}p{1.05cm}}
\hline
Dataset      & CUHK01 & CUHK03 & SYSU  &  Market    & ExMarket & MSMT17    \\
\hline
Measure      & single & single  & single  & multi(MAP) & multi(MAP) & multi(MAP) \\
\hline
DECAMEL$_i$      & 48.33& 28.59 & 18.71  & 40.14(15.89) & 41.06(14.40) & 25.65(8.32)\\
DECAMEL$_r$  &failed & failed &failed &failed &failed &failed \\
DECAMEL      &\textbf{65.81}& \textbf{38.27}  &\textbf{36.14}& \textbf{60.24}(\textbf{32.44}) & \textbf{62.97}(\textbf{33.28}) & \textbf{30.34}(\textbf{11.13}) \\
\hline
\end{tabular}%
\vspace{-0.2cm}
\end{table}

\begin{table}[!t]
\renewcommand{\arraystretch}{1.1}
\caption{Evaluation when given label information to a small proportion of training samples on Market.
Similar observations can be made on other datasets.
``Acc.'' is the rank-1 matching rate measured in \%.}
\vspace{-0.4cm}
\label{table:semi}
\centering
\scriptsize
\begin{tabular}{ccccc}
\hline
Proportion & 0\% & 10\% & 20\% & 30\% \\
\hline
Acc. (MAP) & 60.24(32.44) & 64.55(38.26) & 67.01(40.33) & 69.98(43.20) \\
\hline
\end{tabular}%
\vspace{-0.2cm}
\end{table}

\subsubsection{Benefiting from Extra Labelled Data}
We evaluate a significant property of our framework: the ability to benefit from a little extra labelled data.
This is a natural way to further boost the performance
for an unsupervised application scenario.
We give label information to a small proportion of training samples,
i.e., $10\%$, $20\%$ and $30\%$.
These labelled samples are separated from the unlabelled samples and form extra clusters according to their labels.
We show in Table \ref{table:semi} the results on Market which is very representative and similar observations can be made on other datasets.
We can see that when given a little extra label information,
the accuracy is improved as well as MAP.

\begin{table}[!t]
\renewcommand{\arraystretch}{1.1}
\caption{Evaluation when the training samples size grows on the largest dataset ExMarket.
``\#'' means the number of. ``Acc.'' is the rank-1 matching rate measured in \%.}
\label{table:grow}
\vspace{-0.4cm}
\centering
\scriptsize
\begin{tabular}{>{\centering\arraybackslash}p{2cm}
>{\centering\arraybackslash}p{1.1cm}
>{\centering\arraybackslash}p{1.1cm}
>{\centering\arraybackslash}p{1.1cm}
>{\centering\arraybackslash}p{1.3cm}}
\hline
\# Training samples & 1,000 & 10,000 & 100,000 & 112,351(all) \\
\hline
Acc. (MAP) & 52.52(22.33) & 60.04(30.17) & 62.86(33.21) & 62.98(33.28) \\
\hline
\end{tabular}%
\vspace{-0.2cm}
\end{table}

\subsubsection{Benefiting from More Unlabelled Data}
In typical unsupervised Re-ID scenarios, e.g., public surveillance,
the available data increases with time.
Therefore, it is significant for an unsupervised Re-ID model to benefit from more unlabelled samples.
We evaluate this property by varying the training set size on the largest dataset ExMarket, which uniquely provides over $100,000$ samples.
We show the results in Table \ref{table:grow}.
We can see that when the training set size
grows, the accuracy and MAP are improved significantly.


\re{\subsubsection{Running Time}}

\begin{table}[!t]
\renewcommand{\arraystretch}{1.1}
\caption{Running time on the Market-1501 dataset. All methods are implemented in MATLAB R2017a on a Linux server.
Note that ISR does not need a training procedure.
Training time of DECAMEL includes all steps in Algorithm \ref{AlgDeCamel}.
Testing time is the average time of each probe image searching over the gallery list.}
\label{table:time}
\vspace{-0.4cm}
\centering
\scriptsize
\begin{tabular}{cccc}
\hline
Method & Dic \cite{Dic} & ISR \cite{ISR} & DECAMEL \\
\hline
Training/Testing Time & 35.2h/0.02s & -/98.0s & 5.6h/0.02s  \\
\hline
\end{tabular}%
\vspace{-0.2cm}
\end{table}

\re{
We report the running time of our model on the Market-1501 dataset in Table \ref{table:time}, compared to the most competitive models Dic \cite{Dic} and ISR \cite{ISR}
which together take all the second places in Table \ref{table:comparison}.
A GTX Titan X GPU is used for deep learning in DECAMEL.
We can see that the testing procedure (i.e. usage after deployment) of DECAMEL is efficient.
}

\re{\subsection{Evaluation on the View Clustering}\label{section:eval_VC}}

\begin{table}[!t]
\renewcommand{\arraystretch}{1.1}
\caption{Comparative results in the view-extendable setting.
The performances of Rank-1 accuracy (MAP) are \emph{\textbf{only}} of all the \emph{\textbf{unseen}} views (see the text in Sec. \ref{section:eval_VC}).
}
\vspace{-0.4cm}
\label{table:comparison_VC}
\centering
\scriptsize
\begin{tabular}{ccccc}
\hline
Method & AML & Dic  & DECAMEL$_{VC}$\\
\hline
Market-1501 & 41.57(15.01) & 43.82(18.19)  & 56.50(29.99) \\
MSMT17 & 19.77(5.87) & 21.04(6.00)  & 26.42(8.75) \\
\hline
\end{tabular}%
\vspace{-0.2cm}
\end{table}

\begin{figure}[!t]
\centering
\includegraphics[width=0.6\linewidth, height=0.4\linewidth]{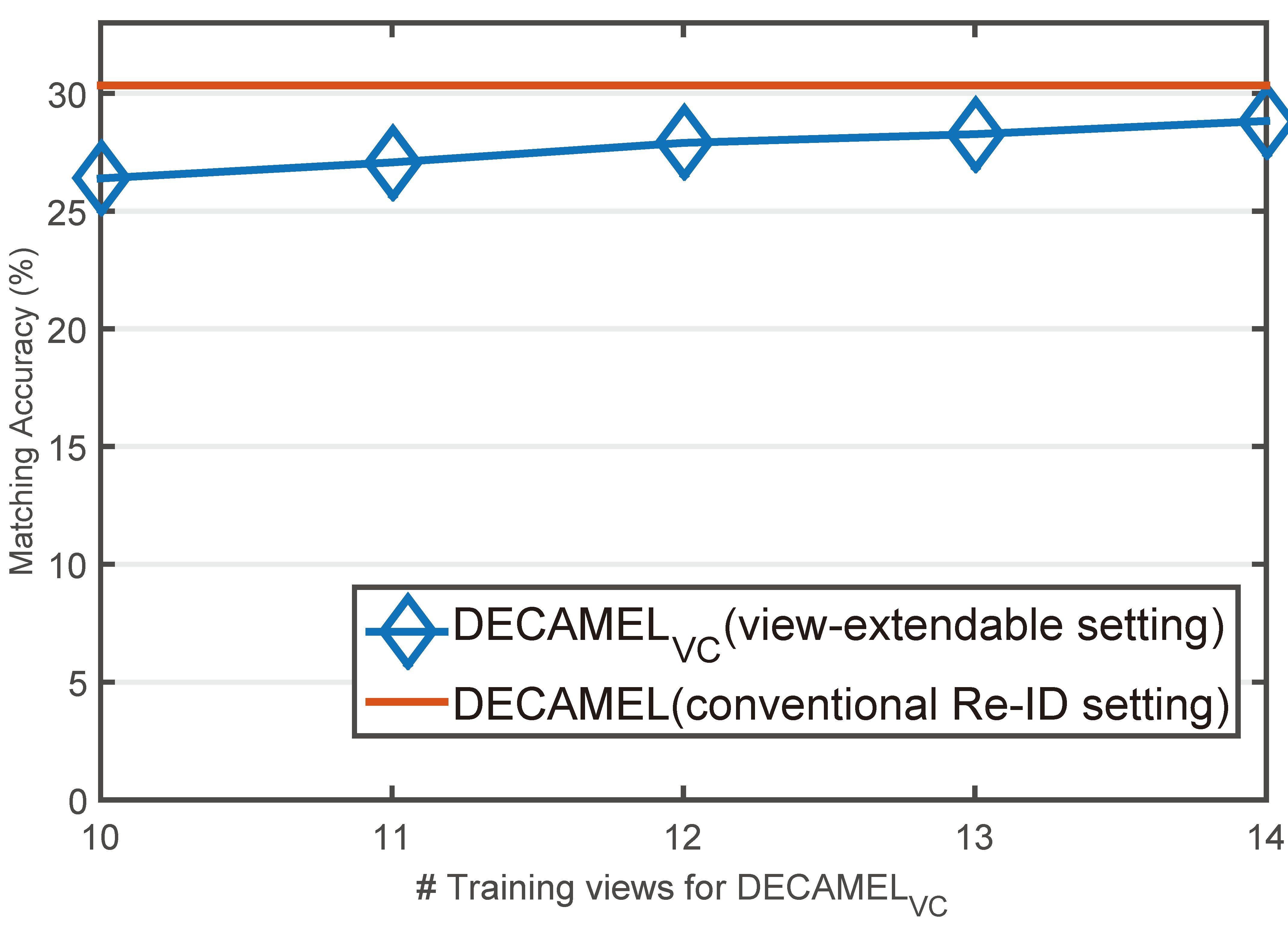}
\vspace{-0.4cm}
\caption{\label{figure:training_view_vary}
Performances of DECAMEL$_{VC}$ in the view-extendable setting
in the MSMT17 dataset. We fix $J=10$.
}
\vspace{-0.4cm}
\end{figure}

%
\re{
In the following we evaluate the View Clustering (VC) proposed in Sec. \ref{section:strategies} for view-extendable
(i.e. requiring to add new views during testing stage) and large-scale applications.
}

\re{
\vspace{0.1cm}
\noindent
\textbf{The View-extendable Re-ID setting}.
To evaluate the performance in the view-extendable scenario instead of the conventional Re-ID setting,
we divide the 15 views of the MSMT17 dataset to two sets: a set of 2/3 of all the views, i.e. 10 views as training views, and a set of the other 1/3, i.e. 5 views as testing views. In the training stage, we only use the training images that are from the training views, but discard all the training images from the testing views. More specifically, we randomly divided all views into 3 subsets each of which contains 1/3 of all the views, and we alternatively select one subset as testing views and the other two as training views, repeat the above procedure for three different subset divisions (i.e. in total we train and test for 3*3=9 times) and report the overall averaged results. In the testing stage, we only use the probe images from the testing views for computing the quantitative results (i.e. Rank-1 accuracy and MAP). In this view-extendable setting, for comparability to the conventional setting, we remain the gallery set to contain images from all views, so that the only difference between these two settings is whether the models have seen training samples from the testing views in the training stage. We also report the results on the Market-1501 dataset on which 2 views are for testing views.
}

\re{
\vspace{0.1cm}
\noindent
\textbf{Comparative results}.
Following the above view-extendable setting, we report the comparative results with the most competitive method (i.e. Dic) \rere{as well as a clustering-based model AML} on the Market-1501 and MSMT17 dataset in Table \ref{table:comparison_VC}.
In our method DECAMEL with View Clustering (denoted as DECAMEL$_{VC}$ and introduced in Sec. \ref{section:strategies}) we set the number of view clusters to $10$ on MSMT17 and $4$ on Market-1501.
}

\re{
From Table \ref{table:comparison_VC}, two observations can be made:
1) Our method also outperforms the compared methods with a clear margin in this view-extendable setting.
2) In the view-extendable setting,
the performances are lower than those in the conventional setting,
but this is reasonable since the training samples from the testing views are not available.
}

\re{
\vspace{0.1cm}
\noindent
\textbf{Generalizability of the View Clustering}.
We further investigate the generalizability of VC when more training views are available.
To this end, we now take $n$ views ($n=5,4,3,2,1$) on the MSMT17 dataset as testing views,
so that we have ($15-n$) training views. In the testing stage, similarly to the above setting, we report the averaged results of the unseen views.
We fix the number of view prototypes $J=10$ and show the results in Figure \ref{figure:training_view_vary}.
}

\re{
From the above results, we can make two observations: 1) When there are more training views, the performance increases. This is because when there are more available training views, the view prototypes can cover a wider range of typical view-specific conditions, and thus more generalizable. 2) the performances of DECAMEL$_{VC}$ are close to DECAMEL, showing that although the testing views are not seen in the training stage, the learned projections of view prototypes can generalize to the unseen views. Note that when the number of training views is 14 DECAMEL with View Clustering (DECAMEL$_{VC}$) still has a gap of $1.5\%$ compared to DECAMEL.
While this observation also indicates that each view has its specific condition, DECAMEL$_{VC}$ thus strikes a balance where the ability to precisely model the view-specific condition is compromised for better generalizability. Moreover, we could infer from the above two observations that in real-world large-scale problems where there would be much more available camera views,
the curve could be further extrapolated and we can expect that the performance of DECAMEL$_{VC}$ could further approximate DECAMEL, i.e. the generalizability of DECAMEL$_{VC}$ in large-scale applications shall be further improved.
}


\vspace{-0.2cm}
\section{Conclusion and Discussion}

In this work, we present a novel approach for unsupervised Re-ID
by formulating it as an unsupervised asymmetric metric learning problem.
We propose a novel unsupervised loss function to produce a deep framework DECAMEL,
which learns the asymmetric metric and embeds it into a deep feature learning network by end-to-end learning.
The experiments show that our model can outperform the related unsupervised Re-ID models.

The analysis and experimental results suggest the effectiveness of the asymmetric modelling in unsupervised Re-ID.
We note that the asymmetric modelling could be extensively embedded into other modelling strategies in unsupervised Re-ID,
e.g., designing view-specific features, learning unsupervised asymmetric metrics and learning view-specific dictionaries.
Our work also suggests the potential of unsupervised metric learning in Re-ID,
especially that based on cross-view clustering.
\re{A future direction could be exploring the behaviour of asymmetric modelling in front of a view-imbalance problem,
where the number of samples in each camera view is largely imbalanced, which could be necessary.
Deriving theoretical guarantees on the robustness against the view-imbalance problem could further perfect the asymmetric modelling theoretically.}
\vspace{-0.5cm}
\section*{Acknowledgment}
This work was supported partially by the National Key Research and Development Program of China (2016YFB1001002), NSFC(61522115, 61661130157, 61472456, U1611461), Guangdong Province Science and Technology Innovation Leading Talents (2016TX03X157), and the Royal Society Newton Advanced Fellowship (NA150459).

This paper has supplementary downloadable material available at http://ieeexplore.ieee.org, provided by the author. The material includes a document of more experiments and analysis. Contact xKoven@gmail.com for further questions about this work.

\bibliographystyle{IEEEtran}
\bibliography{Koven}

\begin{IEEEbiography}
[{\includegraphics[width=1in,height=1.25in,clip,keepaspectratio]{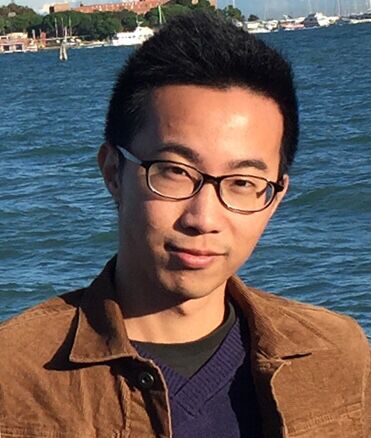}}]
{Hong-Xing Yu}
received the bachelor's degree in communication engineering from Sun Yat-Sen University in 2017.
He is now a M.S. student in the School of Data and Computer Science in Sun Yat-Sen University.
His research interest lies in computer vision and machine learning. \\ Homepage: \url{http://isee.sysu.edu.cn/~yuhx/}.
\end{IEEEbiography}
\begin{IEEEbiography}
[{\includegraphics[width=1in,height=1.25in,clip,keepaspectratio]{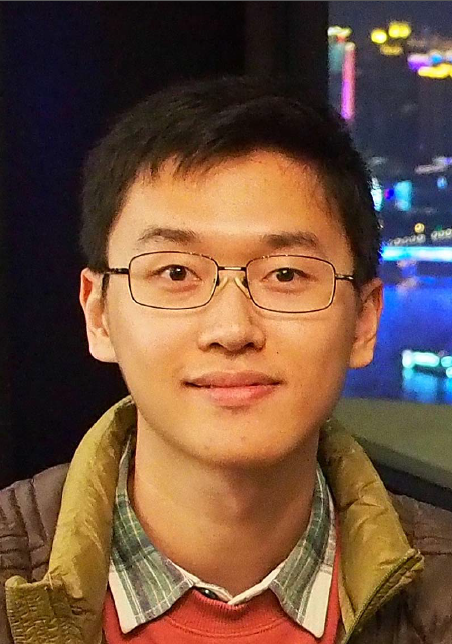}}]
{Ancong Wu}
received the bachelor's degree in intelligence science and technology from Sun Yat-Sen University in 2015.
He is pursuing PhD degree with the School of Electronics and Information Technology in Sun Yat-sen University.
His research interests are computer vision and machine learning. He is currently focusing on the topic of person re-identification.
\\ Homepage:\url{http://isee.sysu.edu.cn/~wuancong/}.
\end{IEEEbiography}
\begin{IEEEbiography}
[{\includegraphics[width=1in,height=1.25in,clip,keepaspectratio]{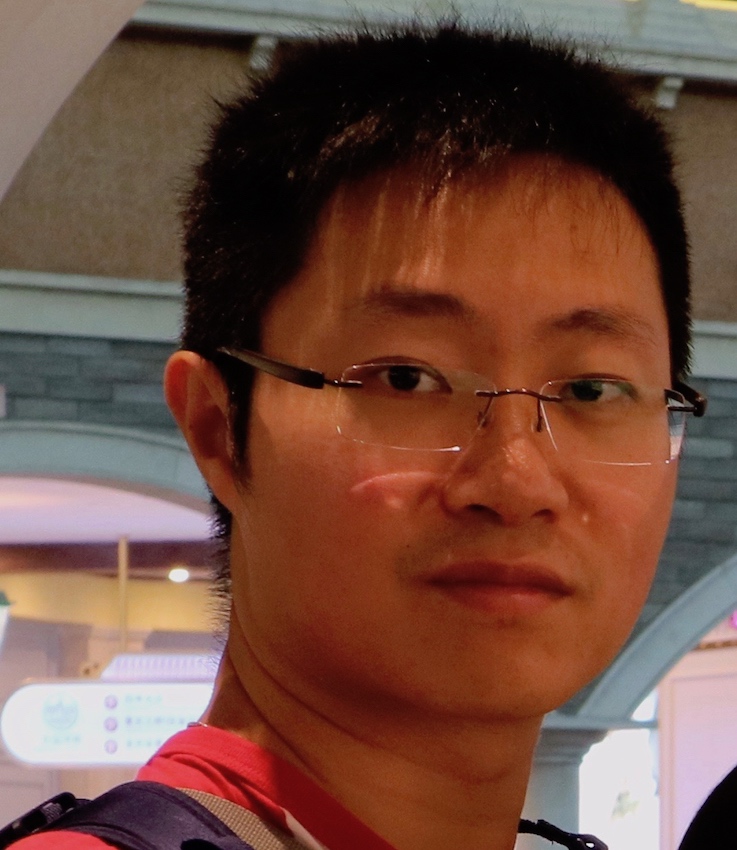}}]
{Wei-Shi Zheng} is now a professor at Sun Yat-sen University. His research interests include person association and activity understanding in visual surveillance. He has now published more than 100 papers, including more than 70 publications in main journals (TPAMI,TNN,TIP,PR) and top conferences (ICCV, CVPR,IJCAI,AAAI). He served as an area chair for AVSS 2012, ICPR 2018 and BMVC 2018, and a Senior PC for IJCAI 2019. He has joined Microsoft Research Asia Young Faculty Visiting Programme. He is a recipient of Excellent Young Scientists Fund of the National Natural Science Foundation of China, and a recipient of Royal Society-Newton Advanced Fellowship, United Kingdom. He is an associate editor of the Pattern Recognition Journal.
\\ Homepage: \url{http://isee.sysu.edu.cn/~zhwshi/}.
\end{IEEEbiography}

\end{document}